%% file: paper.tex
\pdfoutput=1

\documentclass{article}
\usepackage{arxiv,times}

\usepackage[dvipsnames,svgnames]{xcolor}
\usepackage[utf8]{inputenc}
\usepackage[T1]{fontenc}
\usepackage{microtype}
\usepackage{graphicx}
\usepackage{caption}
\usepackage{subcaption}
\usepackage{array}
\usepackage{booktabs}
\usepackage{float}
\usepackage{enumitem}
\usepackage{placeins}
\usepackage{amsmath}
\usepackage{amssymb}
\usepackage{amsfonts}
\usepackage{mathtools}
\usepackage{tikz}
\usetikzlibrary{positioning,fit,backgrounds}
\usepackage{url}
\usepackage{hyperref}
\hypersetup{colorlinks=true,allcolors=NavyBlue,pdftitle={Good Memory has ECC: Evaluating the Memory of Vision-Language Models Beyond Accuracy},pdfauthor={Shmuel Berman, Jia Deng}}
\usepackage{cleveref}
\crefname{appendix}{appendix}{appendices}
\Crefname{appendix}{Appendix}{Appendices}

\usepackage{titlesec}
\titlespacing*{\paragraph}{0pt}{3.25ex plus 1ex minus .2ex}{1em}

\title{Good Memory has ECC: Evaluating the Memory of Vision-Language Models Beyond Accuracy}

\newcommand{\codeurl}{https://github.com/princeton-vl/ECCBench}

\author{%
  Shmuel Berman \\
  Department of Computer Science \\
  Princeton University \\
  \texttt{sb6870@princeton.edu} \\
  \And
  Jia Deng \\
  Department of Computer Science \\
  Princeton University \\
  \texttt{jiadeng@princeton.edu} \\
}

\begin{document}

\maketitle
{\renewcommand{\thefootnote}{}\footnotetext{Preprint. Under review.}}

\input{main.tex}

\bibliography{example_paper}
\bibliographystyle{plainnat}

\newpage
\appendix

\input{appendix.tex}

\end{document}

%% file: main.tex
\begin{abstract}

Memory is widely viewed as an important unsolved problem for LLMs and VLMs, and current benchmarks typically evaluate it by testing accuracy over long text or video. However, accuracy alone misses properties that matter for real long-horizon tasks. We introduce ECCBench, a benchmark and evaluation protocol that measures memory beyond a system's \emph{capacity}---its raw accuracy at a specific budget---via three axes we call ECC: \emph{efficiency}---the computation, in FLOPs, needed to answer from memory; \emph{compression}---whether compressible inputs are remembered more accurately or efficiently; and \emph{calibration}---whether the system abstains in response to its own uncertainty and the cost of an error. We find that pretrained VLMs compress their memory over text but not video and are poorly calibrated on both. Among a broader set of memory backbones, several non-Transformer architectures achieve better compression-calibration tradeoffs than RoPE Transformers, suggesting they may be useful components for agents operating over long horizons.
\end{abstract}%

\section{Introduction}

Memory is widely viewed as an important missing capability for LLMs and VLMs, and is a subject of active research~\citep{du2025mom, long2025m3agent, sinha2026illusion}. But what does it mean for an LLM or VLM to have a good memory and how can we evaluate it? This question may sound trivial and appears to have a settled solution already widely practiced by existing benchmarks~\citep{packer2023memgpt,hu2025memoryagentbench,wang2024lvbench}. The solution is to measure accuracy of answering questions about a long text or video---if a model achieves high accuracy, it has a good memory.

But is it so? In this paper, we argue that it is not---measuring accuracy alone is insufficient and does not constitute proper evaluation of memory. We argue that additional aspects must be measured, which include Efficiency, Compression, and Calibration (ECC).

\begin{figure}[!htbp]
\centering
\includegraphics[width=.82\textwidth,trim={0 9bp 0 1bp},clip]{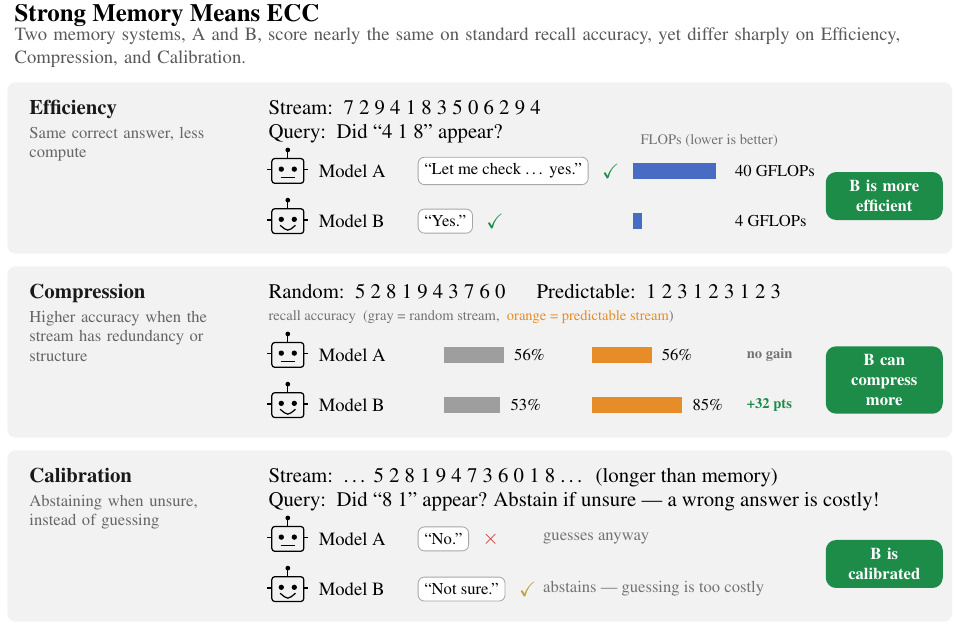}
\caption{\textbf{Accuracy alone cannot tell two memory systems apart, but ECC can.} Two memory systems, A and B, score nearly the same on standard accuracy but differ on the axes we measure beyond accuracy: \emph{Efficiency} (computation spent per query), \emph{Compression} (the accuracy the system gains when the input has redundant structure in it), and \emph{Calibration} (higher abstention rates on questions where it cannot recall the answer and an error is costly). System B is the stronger memory on all three.}
\label{fig:ecc}
\end{figure}

Efficiency---the computational resources needed to access memorized information---is important to measure in addition to accuracy, because without it, strong memory can be trivial. If given unlimited time, a capable but memoryless model can achieve high accuracy by simply brute-force checking every chunk of the original input, like how a person can re-read every page of a book to find the answer. Intuitively, good memory implies fast and effortless recall of information. A proper evaluation of memory must consider compute cost in addition to accuracy.

Now, if a system can accurately and efficiently answer questions about long inputs, does it necessarily have a good memory? Surprisingly not, unless the questions are chosen appropriately, because a system with poor memory can still do well by exploiting biases in the data it sees and the questions it is asked. For example, if faces occur only occasionally in a long video but every question is about them, a system that memorizes faces and nothing else beats one that remembers everything else. What if we make the questions completely random such that every piece of information will be queried with equal likelihood? In this case, given long enough input and finite storage, choosing what to remember will not help, and the only way to outperform others is \emph{compression}---exploiting the redundancy and regularity in the input (e.g., recurring green-yellow-red patterns of a traffic light).

This suggests that proper evaluation of memory should separate compression from \emph{saliency}---priors on what information is more important to retain. Such priors are highly task- and domain-dependent, and can in fact be counterproductive if questions are adversarially chosen. However, existing evaluations have not been designed to separate saliency as a confounder, and it is often unclear whether a system performs well due to saliency or compression.

Memory in real systems is also finite and cannot possibly remember everything even with compression. Thus a useful memory system needs to have \emph{calibration}: it abstains in response to its own uncertainty and the cost of an error. A memory module should be able to abstain from responding when it detects the requested information was not preserved or is uncertain and there is a cost to incorrect retrieval. However, existing benchmarks have not been designed to evaluate calibration because in most cases the models are not provided an abstain option or a cost for an incorrect answer.

We present ECCBench, a \emph{general} evaluation of memory that takes all of these aspects of memory into account. We measure \emph{efficiency} through the computation required to produce an answer. Our dataset lets us measure \emph{compression} by varying the information content of the input streams and comparing relative accuracy. Finally, we measure \emph{calibration} by giving the system an explicit abstention option with an associated cost of error. Together, these dimensions provide a more complete picture of how effectively a system can store, retrieve, and use information. Figure~\ref{fig:ecc} illustrates the three axes. In the task, a system watches a stream and is later asked whether a short candidate appeared in it. Our code is available \href{\codeurl}{here} and the dataset \href{https://huggingface.co/datasets/anonstreammem/substream-recollection/tree/main}{here}.

Our evaluation reveals significant weaknesses in VLMs. While some VLMs compress their memory on text, the ones we tested generally cannot on video. They are also poorly calibrated across both modalities; increasing the cost of an incorrect answer has little effect on their coverage.

Motivated by our VLM results, we evaluate a diverse set of memory backbones beyond the sequence lengths seen during training. We find that several architectures achieve substantially better compression-calibration tradeoffs than RoPE Transformers. Our experiments show that the underlying architecture itself can substantially change the accuracy, compression, and calibration properties of memory, and motivate developing systems with better compressive abilities.

\section{Related Work}

\paragraph{Definitions of memory.}
Neural network memory has been well-studied in recurrent models such as RNNs, which retain a hidden state across time. Beyond these architectures, other research has conceived of memory as an auxiliary state used to preserve information about the past. Earlier work on memory-augmented networks describes a RAM-like memory table which can be written and read from~\citep{weston2014memorynet,graves2014ntm}. Agentic frameworks distinguish episodic memory of past experiences from semantic memory of facts and procedural memory of skills~\citep{sumers2024coala}, and use summarization or context compression to preserve information over time~\citep{packer2023memgpt,long2025m3agent}. Recent work has even treated feed-forward layers~\citep{geva2021ffnmemories},  the model weights, and optimizer states during training~\citep{behrouz2025nested} as associative memories. In practice, most LLMs use a context window composed of a limited number of tokens and extend their memory by increasing its size or retrieving from external storage~\citep{team2024gemini,lewis2020rag,retreivepmlr-v162-borgeaud22a}. We define memory as an empirical property---agnostic of any particular architecture---defined by accuracy, compute, and abstention on different types of data.

\paragraph{Evaluations of memory.}
There is a rich corpus of work that evaluates memory by asking questions about a long document or video on both synthetic and natural data~\citep{bai2024longbench,an2023levalinstitutingstandardizedevaluation,zhang2024inftybenchextendinglongcontext,fu2024videomme,wu2024longvideobenchbenchmarklongcontextinterleaved,lin2024streamingbench,hsieh2024ruler,zhao2025videoniah,wang2024mmniah}. These test memory through accuracy but confound memory with saliency, or the ability to predict what will be asked, and do not measure other aspects of memory, which become important as the input stream becomes very long. HELM is the closest precedent to our work, but it tests only textual tasks and does not measure compression~\citep{liang2023helm}. MemBench and LongMemEval also evaluate memory beyond accuracy but their benchmarks require conversational reasoning, making it difficult to ascribe failures to poor memory~\citep{tan2025membench,wu2025longmemeval}. In ECCBench, we evaluate the core aspects of memory and disentangle other skills by using associative recall, a standard test of memory~\citep{graves2014ntm,arora2024zoology} that requires little world knowledge or reasoning.

\paragraph{Compression.}
In prior work, compression usually refers to a reduction in storage space by exploiting redundancy in data~\citep{deletang2024lmcompression,rae2019compressive,xiao2023streamingllm,li2024snapkv}. We instead define compression as an increase in accuracy or efficiency---defined through FLOPs---achieved by exploiting redundancy in data. Our definition specifically excludes using knowledge of the task to decide what to retain, which we call \emph{saliency}. \citet{mangalam2023egoschema} formalize one conception of saliency with their \emph{temporal certificate}, the smallest portion of a video which is necessary to verify an answer. Instead of choosing which data to discard, compression uses regularity to store information in a succinct way so it can be accessed more easily. Prior work observed that repetition improves recall in recurrent models~\citep{arora2024justreadtwice}, but has not generalized this to data with predictable structure.

\paragraph{Calibration and abstention.}
Previous research on calibration has tested whether a model's numerical confidence predicts its correctness, and has studied optimal abstention rates~\citep{guo2017calibration,elyaniv2010foundations}. We instead focus on empirical calibration by giving the model different penalty structures and evaluating how often it abstains. This approach is also used in recent work such as \citet{wu2025answerrefuse} and \citet{mohamadi2025hesitation}, who investigate if models are calibrated to the cost of errors on reasoning and knowledge tasks.

\section{Background}\label{sec:background}
\label{sec:eval_method}

We first define memory as an online problem over streams of potentially unbounded length, and specify the properties that we evaluate. A memory system observes an input, keeps some record of it, and later answers
questions using the state it kept. The input may arrive piecemeal and a
question may be asked at any timestep. We write each observation---for example, a video frame---as $x_i$ and the entire stream as $x_{1:L}$. $L$ is not revealed to the system in advance, and we focus on long-horizon memory and the performance as $L$ grows. A memory system must contain a state and provide write and read operations. A write incorporates the next observation
into the state and a read takes a question $q$ and answers based on the state:
\begin{equation}
S_{t+1} = \mathsf{Write}(S_t, x_{t+1}),
\qquad
(\hat y, S_t^{+}) = \mathsf{Read}(S_t, q),
\qquad
\hat y \in \mathcal Y \cup \{\bot\}.
\end{equation}
Here $\mathcal Y$ is the set of allowed answers and $\bot$ is abstention. A write is a general operation that
may append, evict, or reorganize the memory state, and a read may occur at any $t$ and return an
updated state.

We treat memory as a black box, except insofar as we have access to the amount of FLOPs it uses during computation: $F^{\mathrm w}$ for reading the stream and
$F^{\mathrm r}$ for answering. For instance, this corresponds in pretrained VLMs to prefill and decode. In this paper, we use the total $F = F^{\mathrm w} + F^{\mathrm r}$, since a memory system
may spend early so that answering is cheap or defer most computation until a question
arrives. We make no assumption on which tactic is superior because this is task-dependent.

Let $\mathcal D$ be a distribution over input streams and $\mathcal Q$ a distribution over queries about them. For streams drawn from $\mathcal D$ and queries drawn from $\mathcal Q$, let $C$ be the probability of correct retrieval from memory and $R$ be the probability that the memory system responds. The probability of an incorrect answer is $R-C$, and the probability of abstention is $1-R$.

Under this setting good memory has four desirable attributes. The first,
capacity, is measured in accuracy; the three we add are the E, C, and C of the title.

\paragraph{Capacity.}
A system has high capacity if it can accurately recall a large amount of incompressible input. For a stream length $L$ and FLOPs budget $F$, we measure capacity as the correct answer rate $C$ on a high-entropy distribution $\mathcal D_{\text{high}}$. Comparing $C$ across stream lengths at fixed $F$ shows how much incompressible input the system can remember. We assume that the distribution $\mathcal D_{\text{high}}$ has no useful redundancy the model can exploit; in our synthetic examples, we define it as a uniform distribution over the vocabulary. Capacity under an unlimited FLOPs budget is analogous to raw accuracy in other benchmarks.

\paragraph{Efficiency.}
A system is efficient if it reaches a given correct answer rate at low cost, where we define cost through computation. We use total FLOPs $F$ as the cost metric and measure efficiency as the smallest $F$ needed to reach a target $C$. We use FLOPs rather than state size or latency, since storage (outside of specialized hardware such as GPUs) is cheap, and latency depends heavily on hardware and implementation~\citep{gu2024mamba,dao2023flashattention2fasterattentionbetter}. In other settings, it may be more useful to use other metrics such as latency or the depth of computation, but using FLOPs is general and lets us consistently compare different architectures. All of our numbers assume one query per stream. 

\paragraph{Compression.}
A system can compress if it uses stream regularity to improve recall or be more efficient. For example, it may learn the pattern of a regular sequence and be better at remembering events than over an incompressible sequence. At fixed $L$, $F$, and $\mathcal Q$, the accuracy gained over incompressible input is
\begin{equation}
\operatorname{Comp}(L,F;\mathcal D)=C_{\mathcal D}(L,F)-C_{\mathcal D_{\text{high}}}(L,F),
\end{equation}
where $C_{\mathcal D}(L,F)$ is the correct answer rate on streams from $\mathcal D$. Alternatively, it can be measured as FLOPs saved when reaching the same $C$, although this is usually more difficult to measure.

On our streams, we measure regularity using a Lempel--Ziv compression-based estimate of entropy $\hat h_{\mathrm{LZ}}$ in bits per symbol~\citep{kontoyiannis1998nonparametric}. Appendix~\ref{app:entropy_generator} contains the exact procedure we use for calculating the entropy of each stream.

We distinguish compression from \emph{saliency}, which we claim is a task specific property. Saliency exploits knowledge of what questions are likely to be asked to decide which information is worth retaining. Compression instead exploits regularity in the stream so that information that is retained can be stored and retrieved more efficiently. Consider the stream distribution $\mathcal D$ and question distribution $\mathcal Q$. For a stream $x$, let $\mathcal K_{\mathcal D,\mathcal Q}(x)$ denote the information that must be retained to answer any question in $\mathcal Q$. Two streams $x$ and $x'$ in $\mathcal D$ have the same $\mathcal K_{\mathcal D,\mathcal Q}(x)$ when
\[
y(x,q)=y(x',q)
\qquad
\text{for every question }q\text{ in }\mathcal Q,
\]
where $y(x,q)$ is the correct answer to $q$ on stream $x$. Thus, $\mathcal K_{\mathcal D,\mathcal Q}(x)$ is exactly the information in the stream that must be retained to answer all future questions correctly. 

This definition lets us disentangle saliency from compression capability. If the questions in $\mathcal Q$ concern only narrow parts of the stream, many differences between streams never matter, and a system can discard them. If a system has limited space and some questions are more likely, it can further prioritize storing the information that those questions require. Compression capability instead exploits regularity in $\mathcal D$ to reduce how much information must be stored in order to preserve the distinctions that do matter. On natural tasks, $\mathcal D$, $\mathcal Q$, and therefore $\mathcal K$ are generally unknown. We define a synthetic task where we can control them explicitly, together with the entropy of $\mathcal D$. This allows us to test compression without worrying about the effects of saliency.

\paragraph{Calibration.}
A system is calibrated if it answers only when its memory is reliable enough relative to the cost of making an error. We define calibration in terms of empirical selectivity and not through a system's stated confidence.

Because the optimal abstention rate depends on the cost of an error, we measure calibration through selectivity and responsiveness. To do so, we first evaluate coverage, $R$, and accuracy-on-answered, $C/R$, on different payoff schemes~\citep{elyaniv2010foundations}. If we define $R_\lambda$ as coverage when a wrong answer costs $\lambda$, answering is
worthwhile when the system's confidence exceeds $\tau = \lambda/(1+\lambda)$,
so raising $\lambda$ should lower coverage.
\emph{Responsiveness} is the drop in coverage from a setting with a low cost of error to one with a high cost of error, $R_{\lambda_{\mathrm{lo}}} -
R_{\lambda_{\mathrm{hi}}}$; \emph{Selectivity} is the difference in accuracy between the questions that are kept and those that were dropped when the cost of error increased.

\section{Experiments}\label{sec:substream}
We use our dataset, ECCBench, to investigate two underexplored aspects of memory: compression and calibration. ECCBench has a synthetic subset, composed of text and video streams with known entropy, and a natural subset composed of natural videos. We evaluate nine pretrained VLM families and a RAG-like pipeline~\citep{lewis2020rag} on ECCBench, capturing the FLOPs used by each model under the protocol in Section~\ref{sec:eval_method} based on the number of generated tokens. Our procedure is also fully compatible with other memory-based systems such as agentic or external memory~\citep{hu2025memoryagentbench,long2025m3agent,graves2014ntm}. The full details of our experimental protocol are in Appendix~\ref{app:pretrained_eval_details}, and FLOPs are calculated via the procedure in Appendix~\ref{app:flops_per_model}.

\subsection{Data}\label{sec:data}

\definecolor{hit}{HTML}{F2A03D}

\begin{figure}[!htbp]
\centering
\resizebox{0.933\textwidth}{!}{\begin{tikzpicture}[
  ttl/.style={font=\scriptsize\bfseries, anchor=west},
  gut/.style={font=\scriptsize\itshape, anchor=east, text=black!55},
]
\def\FS{1.14}                       %
\def\NW{1.86}                       %
\def\G{1.44}                        %
\def\C{1.56}                        %
\def\PB{1.30}                       %
\def\PN{1.90}                       %

\node[ttl] at (0,0) {(a) Synthetic text};
\node[gut] at (\G,-0.40) {stream};
\node[font=\scriptsize\ttfamily, anchor=west] at (\C,-0.40)
  {8, 1, \fcolorbox{hit!90!black}{hit!30}{4, 6, 2}, 1, 1, 2, 13, 1, 2, 9, 3, 1, 2, 7, 0, 5, 2, \dots};
\node[gut] at (\G,-0.86) {candidate};
\node[font=\scriptsize\ttfamily, anchor=west] at (\C,-0.86)
  {\fcolorbox{hit!90!black}{hit!30}{4, 6, 2}};

\begin{scope}[yshift=-1.30cm]
\node[ttl] at (0,0) {(b) Synthetic video};
\node[gut] at (\G,-0.76) {stream};
  \pgfmathsetmacro{\xx}{\C+0.5*\FS+(1-1)*\PB}
  \node[inner sep=0pt] at (\xx,-0.76) {\includegraphics[width=\FS cm]{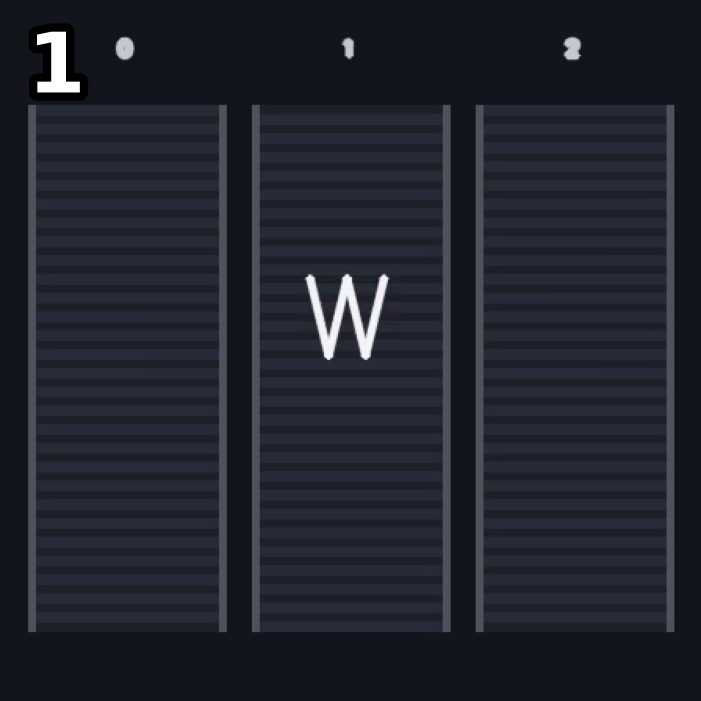}};
  \draw[hit, line width=0.9pt, rounded corners=1.3pt]
     (\xx-0.5*\FS-0.04,-0.76-0.5*\FS-0.04) rectangle (\xx+0.5*\FS+0.04,-0.76+0.5*\FS+0.04);
  \pgfmathsetmacro{\xx}{\C+0.5*\FS+(2-1)*\PB}
  \node[inner sep=0pt] at (\xx,-0.76) {\includegraphics[width=\FS cm]{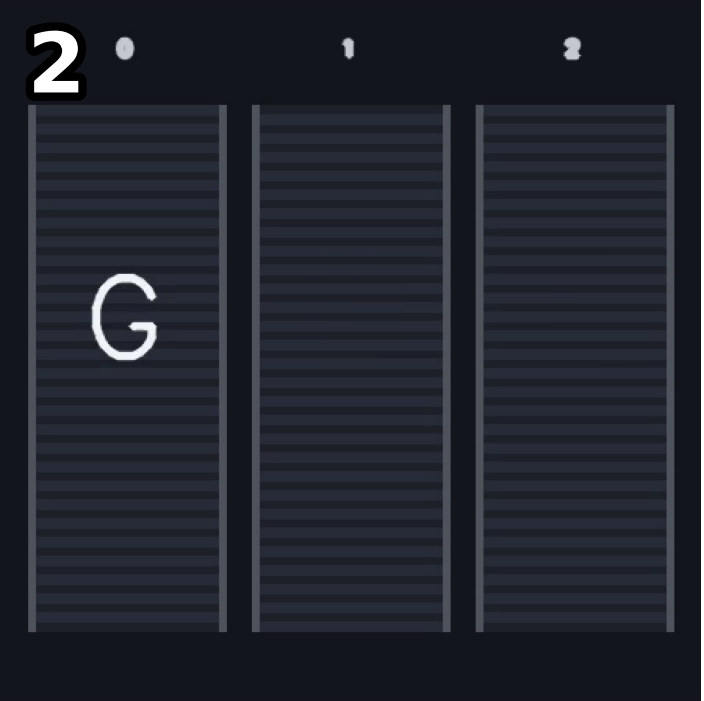}};
  \draw[hit, line width=0.9pt, rounded corners=1.3pt]
     (\xx-0.5*\FS-0.04,-0.76-0.5*\FS-0.04) rectangle (\xx+0.5*\FS+0.04,-0.76+0.5*\FS+0.04);
  \pgfmathsetmacro{\xx}{\C+0.5*\FS+(3-1)*\PB}
  \node[inner sep=0pt] at (\xx,-0.76) {\includegraphics[width=\FS cm]{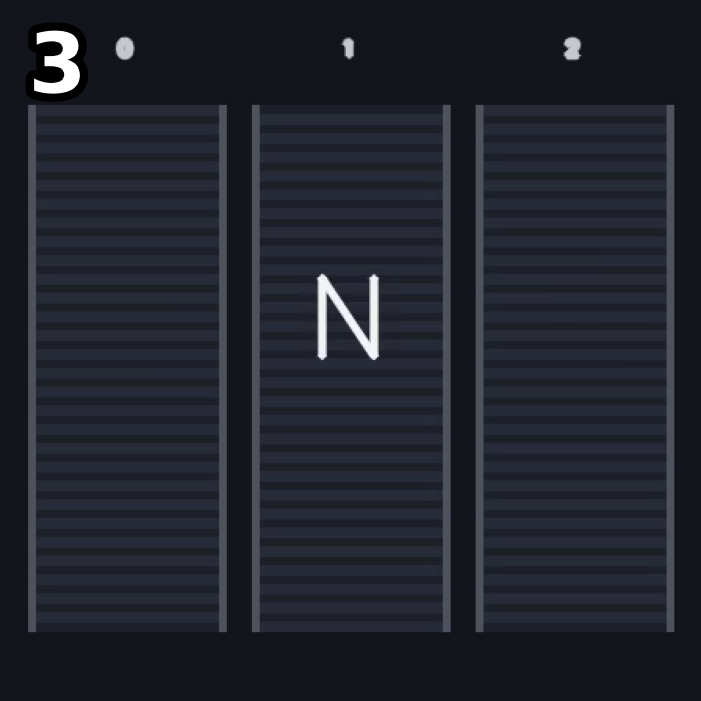}};
  \draw[hit, line width=0.9pt, rounded corners=1.3pt]
     (\xx-0.5*\FS-0.04,-0.76-0.5*\FS-0.04) rectangle (\xx+0.5*\FS+0.04,-0.76+0.5*\FS+0.04);
  \pgfmathsetmacro{\xx}{\C+0.5*\FS+(4-1)*\PB}
  \node[inner sep=0pt] at (\xx,-0.76) {\includegraphics[width=\FS cm]{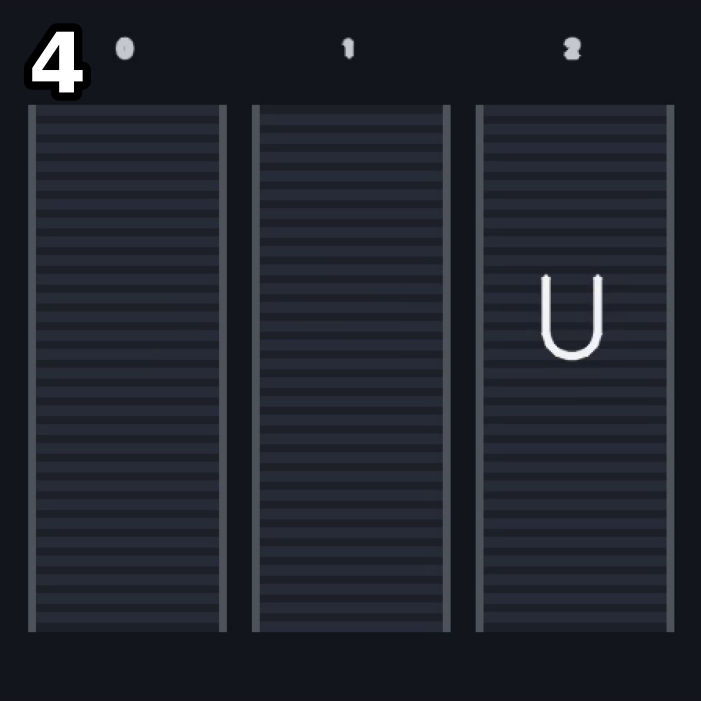}};
  \draw[hit, line width=0.9pt, rounded corners=1.3pt]
     (\xx-0.5*\FS-0.04,-0.76-0.5*\FS-0.04) rectangle (\xx+0.5*\FS+0.04,-0.76+0.5*\FS+0.04);
  \pgfmathsetmacro{\xx}{\C+0.5*\FS+(5-1)*\PB}
  \node[inner sep=0pt] at (\xx,-0.76) {\includegraphics[width=\FS cm]{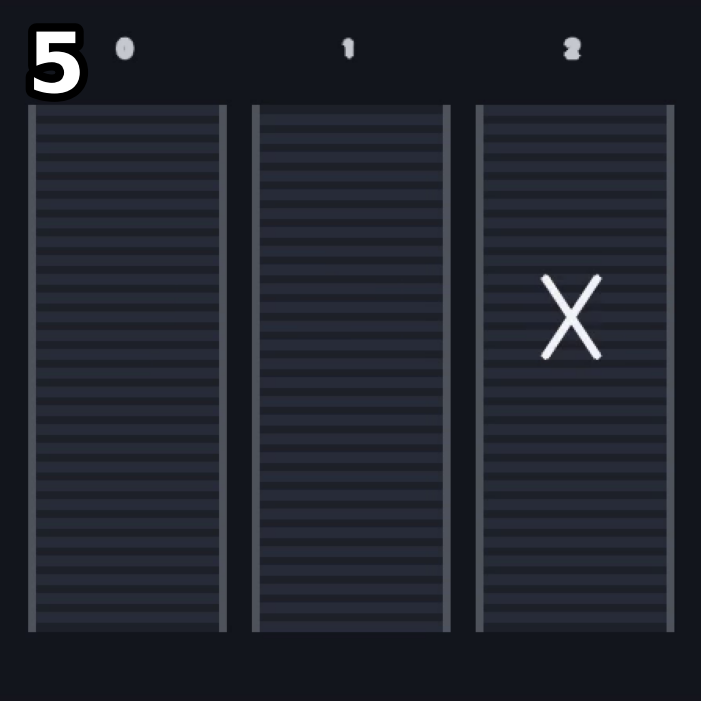}};
  \draw[hit, line width=0.9pt, rounded corners=1.3pt]
     (\xx-0.5*\FS-0.04,-0.76-0.5*\FS-0.04) rectangle (\xx+0.5*\FS+0.04,-0.76+0.5*\FS+0.04);
  \pgfmathsetmacro{\xx}{\C+0.5*\FS+(6-1)*\PB}
  \node[inner sep=0pt] at (\xx,-0.76) {\includegraphics[width=\FS cm]{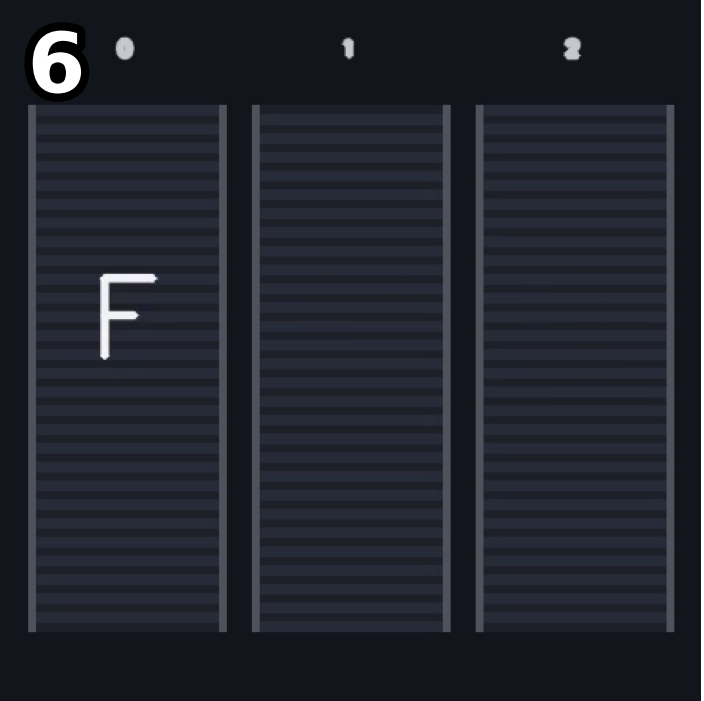}};
  \pgfmathsetmacro{\xx}{\C+0.5*\FS+(7-1)*\PB}
  \node[inner sep=0pt] at (\xx,-0.76) {\includegraphics[width=\FS cm]{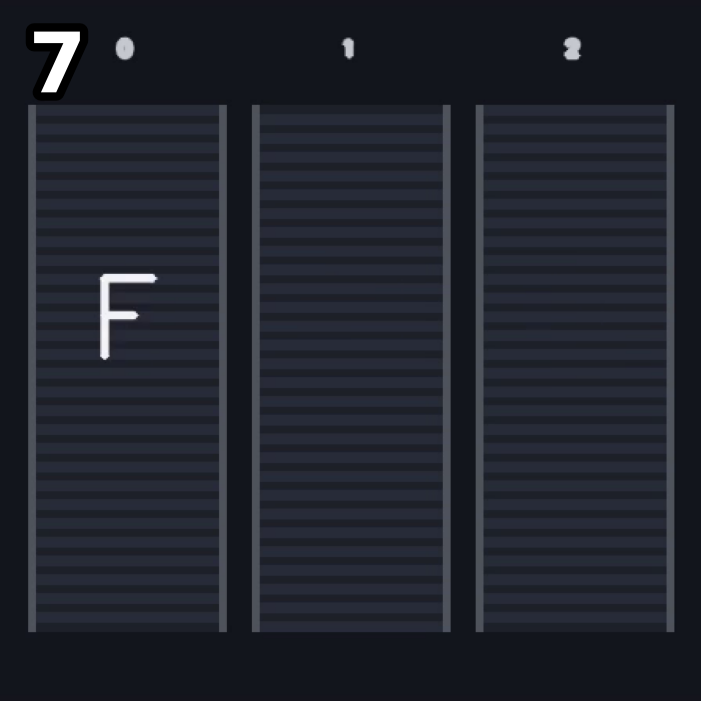}};
  \pgfmathsetmacro{\xx}{\C+0.5*\FS+(8-1)*\PB}
  \node[inner sep=0pt] at (\xx,-0.76) {\includegraphics[width=\FS cm]{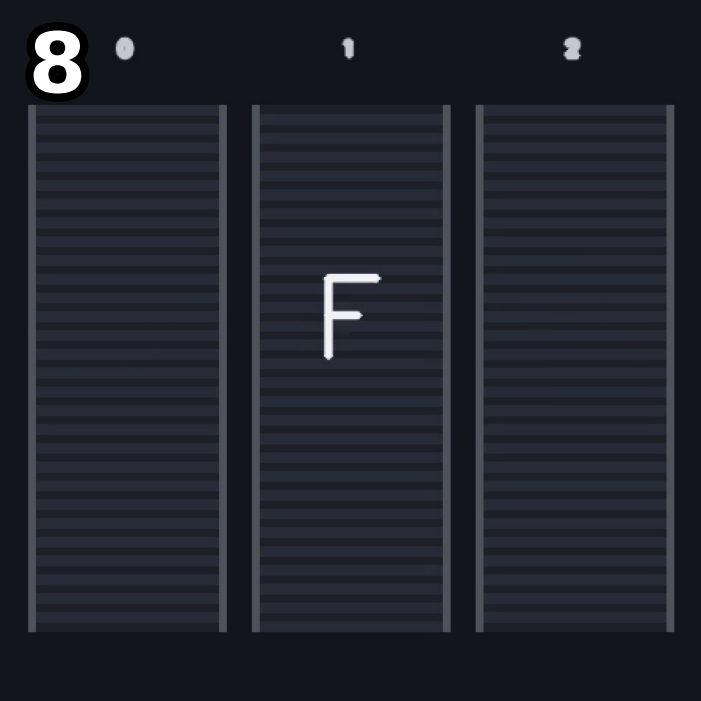}};
\node[font=\scriptsize] at (\C+0.5*\FS+8*\PB-0.16,-0.76) {$\cdots$};
\node[gut] at (\G,-2.02) {candidate};
  \pgfmathsetmacro{\xx}{\C+0.5*\FS+(1-1)*\PB}
  \node[inner sep=0pt] at (\xx,-2.02) {\includegraphics[width=\FS cm]{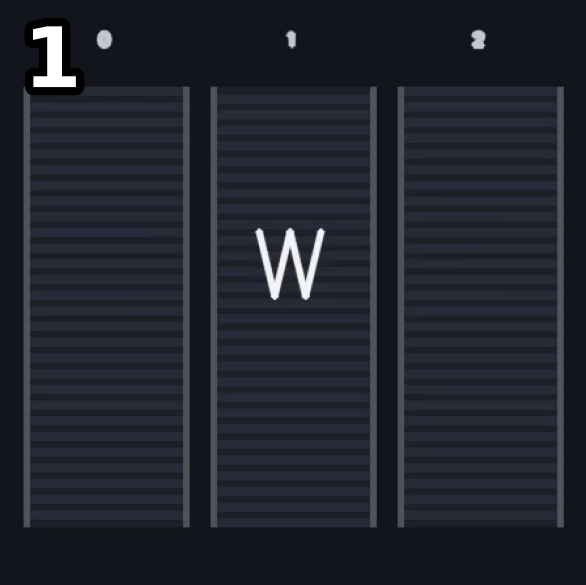}};
  \draw[hit, line width=0.9pt, rounded corners=1.3pt]
     (\xx-0.5*\FS-0.04,-2.02-0.5*\FS-0.04) rectangle (\xx+0.5*\FS+0.04,-2.02+0.5*\FS+0.04);
  \pgfmathsetmacro{\xx}{\C+0.5*\FS+(2-1)*\PB}
  \node[inner sep=0pt] at (\xx,-2.02) {\includegraphics[width=\FS cm]{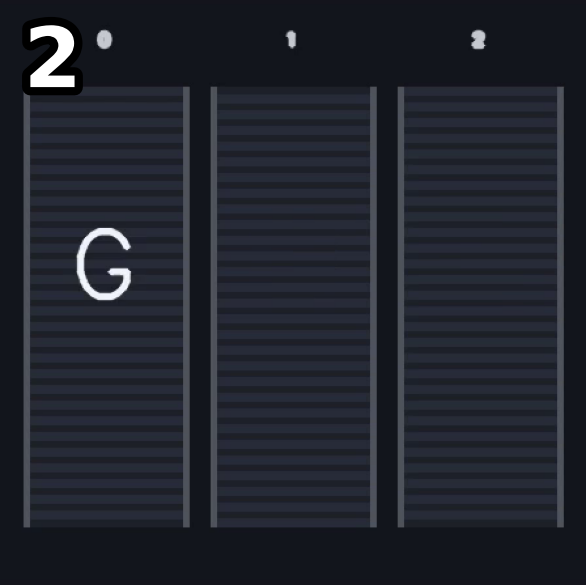}};
  \draw[hit, line width=0.9pt, rounded corners=1.3pt]
     (\xx-0.5*\FS-0.04,-2.02-0.5*\FS-0.04) rectangle (\xx+0.5*\FS+0.04,-2.02+0.5*\FS+0.04);
  \pgfmathsetmacro{\xx}{\C+0.5*\FS+(3-1)*\PB}
  \node[inner sep=0pt] at (\xx,-2.02) {\includegraphics[width=\FS cm]{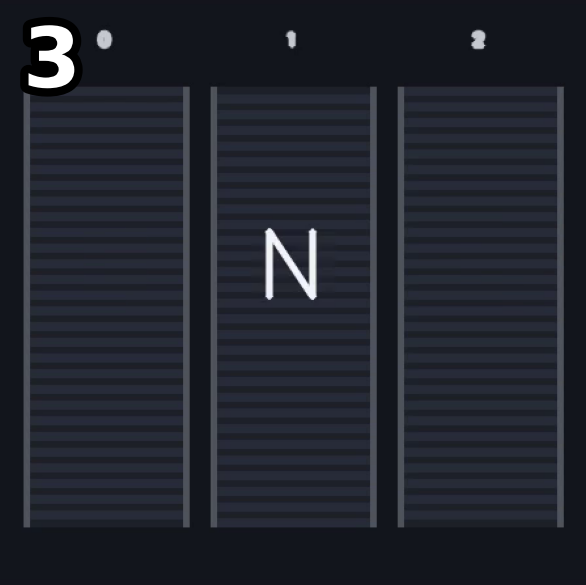}};
  \draw[hit, line width=0.9pt, rounded corners=1.3pt]
     (\xx-0.5*\FS-0.04,-2.02-0.5*\FS-0.04) rectangle (\xx+0.5*\FS+0.04,-2.02+0.5*\FS+0.04);
  \pgfmathsetmacro{\xx}{\C+0.5*\FS+(4-1)*\PB}
  \node[inner sep=0pt] at (\xx,-2.02) {\includegraphics[width=\FS cm]{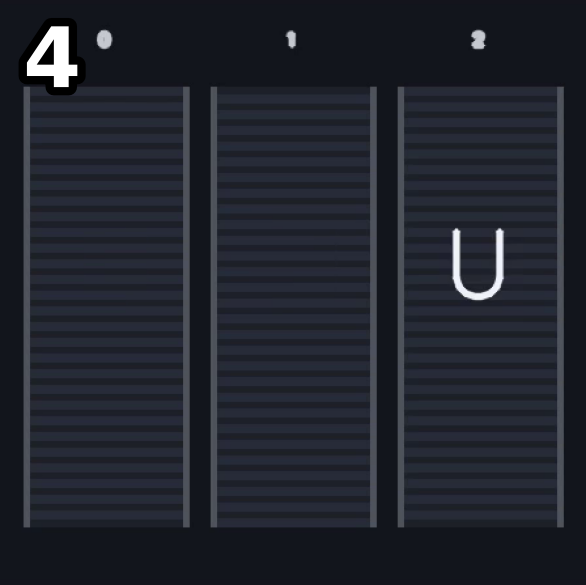}};
  \draw[hit, line width=0.9pt, rounded corners=1.3pt]
     (\xx-0.5*\FS-0.04,-2.02-0.5*\FS-0.04) rectangle (\xx+0.5*\FS+0.04,-2.02+0.5*\FS+0.04);
  \pgfmathsetmacro{\xx}{\C+0.5*\FS+(5-1)*\PB}
  \node[inner sep=0pt] at (\xx,-2.02) {\includegraphics[width=\FS cm]{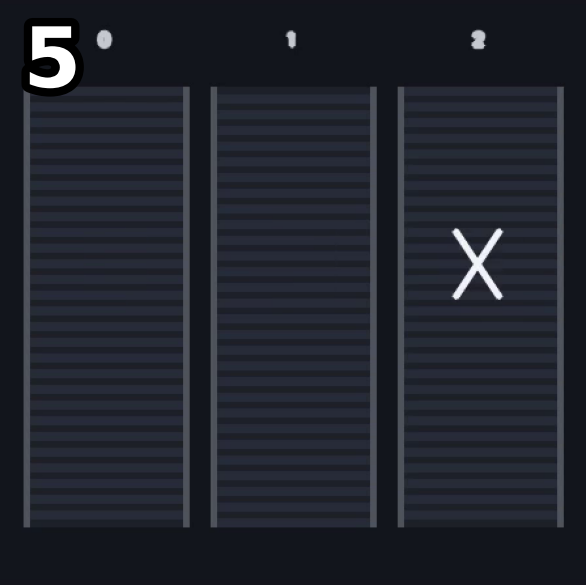}};
  \draw[hit, line width=0.9pt, rounded corners=1.3pt]
     (\xx-0.5*\FS-0.04,-2.02-0.5*\FS-0.04) rectangle (\xx+0.5*\FS+0.04,-2.02+0.5*\FS+0.04);
\end{scope}

\begin{scope}[yshift=-4.14cm]
\node[ttl] at (0,0) {(c) Natural video};
\node[gut] at (\G,-0.70) {stream};
  \pgfmathsetmacro{\xx}{\C+0.5*\NW+(1-1)*\PN}
  \node[inner sep=0pt] at (\xx,-0.70) {\includegraphics[width=\NW cm]{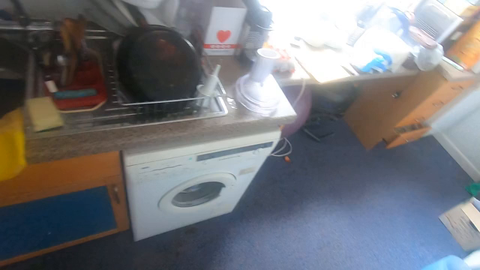}};
  \pgfmathsetmacro{\xx}{\C+0.5*\NW+(2-1)*\PN}
  \node[inner sep=0pt] at (\xx,-0.70) {\includegraphics[width=\NW cm]{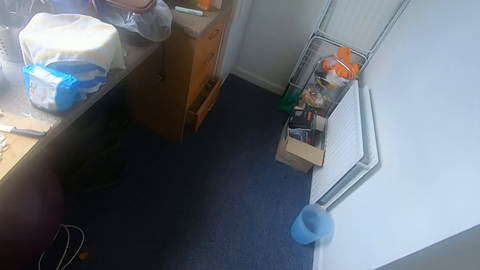}};
  \pgfmathsetmacro{\xx}{\C+0.5*\NW+(3-1)*\PN}
  \node[inner sep=0pt] at (\xx,-0.70) {\includegraphics[width=\NW cm]{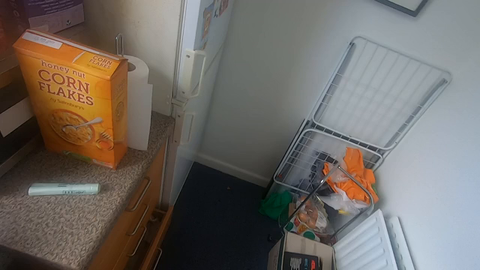}};
  \pgfmathsetmacro{\xx}{\C+0.5*\NW+(4-1)*\PN}
  \node[inner sep=0pt] at (\xx,-0.70) {\includegraphics[width=\NW cm]{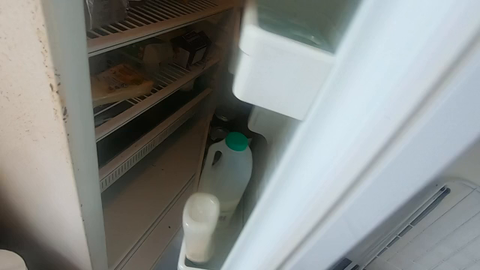}};
  \pgfmathsetmacro{\xx}{\C+0.5*\NW+(5-1)*\PN}
  \node[inner sep=0pt] at (\xx,-0.70) {\includegraphics[width=\NW cm]{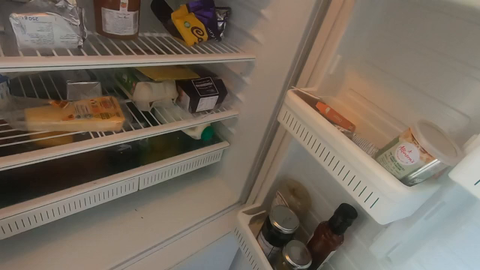}};
  \pgfmathsetmacro{\xx}{\C+0.5*\NW+(6-1)*\PN}
  \node[inner sep=0pt] at (\xx,-0.70) {\includegraphics[width=\NW cm]{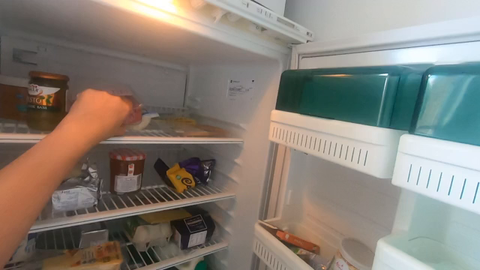}};
\node[font=\scriptsize] at (\C+0.5*\NW+6*\PN-0.42,-0.70) {$\cdots$};
\node[gut] at (\G,-1.42) {candidate};
\node[draw=black!50, rounded corners=1pt, fill=black!7, inner sep=3pt,
      font=\scriptsize, anchor=west] at (\C,-1.42)
      {``Did the person open the fridge in this video?''};
\end{scope}
\end{tikzpicture}}
\caption{\textbf{An example from each subset of ECCBench.} Each example is composed of a long stream and a candidate substream. The model judges whether the candidate occurred in the
stream. On both synthetic subsets the candidate is in the same modality as the
stream---text in (a), video in
(b)--- and for the natural subset (c) the stream is video and the candidate a text
question about an event in the video. Frames in (c) are EPIC-Kitchens-100 (CC BY-NC 4.0).}
\label{fig:data-examples}
\end{figure}

\paragraph{Synthetic subset.}
The streams in our synthetic subset are sampled over a vocabulary $\mathcal V$, with $|\mathcal V|=16$. For the high-entropy distribution $\mathcal D_{\text{high}}$, each symbol is sampled i.i.d.\ and uniformly from $\mathcal V$. Low-entropy streams are generated using variable-order $n$-gram processes at differing levels of entropy. The query distribution $\mathcal Q$ is roughly balanced between true positives and true negatives, where positives are directly copied from the original stream. Each stream has 16 queries. Because of our construction, $\mathcal K_{\mathcal D,\mathcal Q}(x)$ is known precisely: the questions range over all sub-streams of lengths three to six, and the answers depend only on the symbol and its location, so that is all a system needs to retain (and not other details, like the background image on video). The full details of data and question generation are in Appendix~\ref{app:entropy_generator}.

Text streams are comma-separated integers and video streams render one letter per frame on one of three lanes. Rendering details are in Appendix~\ref{app:video_rendering}. Figure~\ref{fig:data-examples} shows one example from each subset.

\paragraph{Natural video subset.}
We build $\mathcal D_{\mathrm{nat}}$ from EPIC-Kitchens-100~\citep{damen2022epic100} and SoccerNet~\citep{deliege2021soccernetv2} clips and create $\mathcal Q_{\mathrm{nat}}$ from activity annotations. Importantly, we can make only qualitative claims about compression on natural video because for natural video we do not know $\mathcal D$ or $\mathcal K$. Additionally, an apparent advantage on low-entropy video can come from compression or, as stated in Section~\ref{sec:background}, from a useful saliency prior. We estimate information content through video encoding and reconstruction at a fixed bandwidth and then split them into low- and high-entropy bands (full details in Appendices~\ref{app:natural_eval} and~\ref{app:natural_density}). Though qualitative, we use natural video to confirm if our results on synthetic data extend to videos that models are more likely to encounter.

\subsection{Can VLMs compress?}\label{sec:compression}
We first ask whether VLMs can compress their memory on text and video. On each of our subsets, we measure \textit{capacity} on the highest entropy streams and \textit{compression} via the difference in accuracy between high- and low-entropy streams. We plot both against efficiency, or FLOPs used, and abstentions count as incorrect. The full details of this evaluation are in Appendix~\ref{app:pretrained_eval_details}.

We do not evaluate any closed-source models on our dataset because we view efficiency as a key part of memory and we cannot calculate how much computation they use. However, we do evaluate them on an even more compressible subset and replicate our main findings in Appendix~\ref{sec:easyhuman}.

\begin{figure}[!htb]
    \centering
    \includegraphics[width=0.875\textwidth,trim={35.28bp 9.960bp 73.08bp 17.160bp},clip]{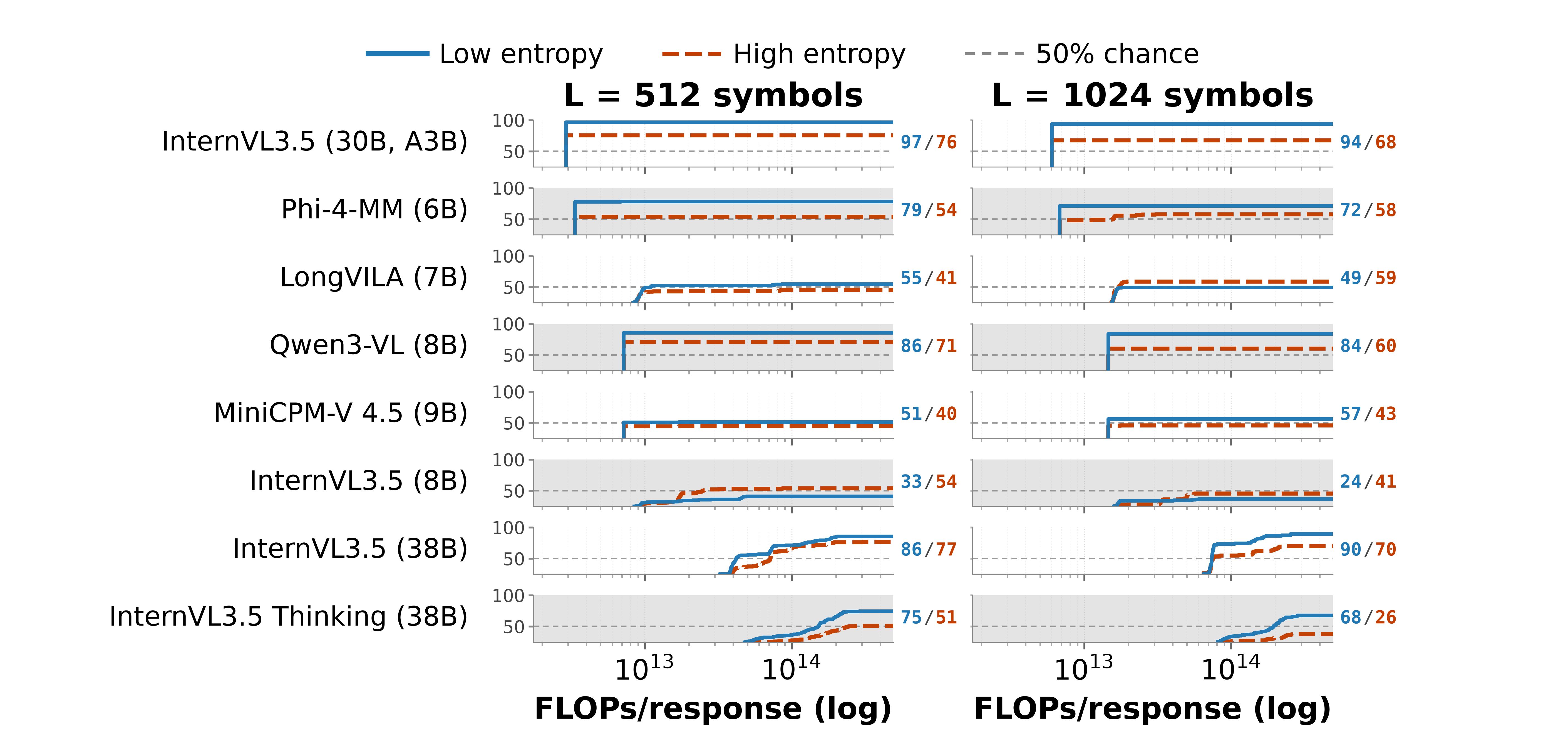}
    \caption{\textbf{On synthetic text, several models compress their memories and a few have high capacity.} Accuracy vs.\ FLOPs at $L{=}512$ (left) and $L{=}1024$ (right) symbols. Blue solid: low-entropy; orange dashed: high-entropy. We define compression capability as the gap above the orange line and below the blue line at a given FLOPs budget; when the orange line is higher, this represents negative compression. Right-edge integers: final accuracy per condition under unlimited FLOPs.}
    \label{fig:flops-1024-pretrained}

\end{figure}

\subsubsection{Text}
\paragraph{VLMs have low capacity on text and compress their memories.} The results of our experiments on text are in Figure~\ref{fig:flops-1024-pretrained}. On low-entropy text, some VLMs retrieve highly accurately. However, their performance drops precipitously on the high-entropy sequences, with only three VLMs scoring significantly above chance, which we define throughout as at least $60\,\%$ accuracy (Appendix~\ref{app:synth_numbers}). We observe that models with high capacity can also generally compress, though the reverse is not true.

\paragraph{VLMs compress by increasing accuracy, not efficiency.} There are two ways to take advantage of regularity: higher accuracy using the same amount of computation, or the same accuracy for less compute. For VLMs, the same accuracy for less compute corresponds to shorter reasoning traces on low-entropy data, so only models with lengthy chain of thought can compress this way. Of the three models that score well on both entropy splits, only InternVL3.5 (38B) reaches similar accuracy at fewer FLOPs on the low-entropy split.  Other models fail to match their 60\,\%+ accuracy on regular data when evaluated on irregular data, regardless of their FLOPs budget.

\begin{figure}[!htb]
    \centering
    \includegraphics[width=0.7459\textwidth,trim={16.32bp 9.960bp 73.08bp 17.160bp},clip]{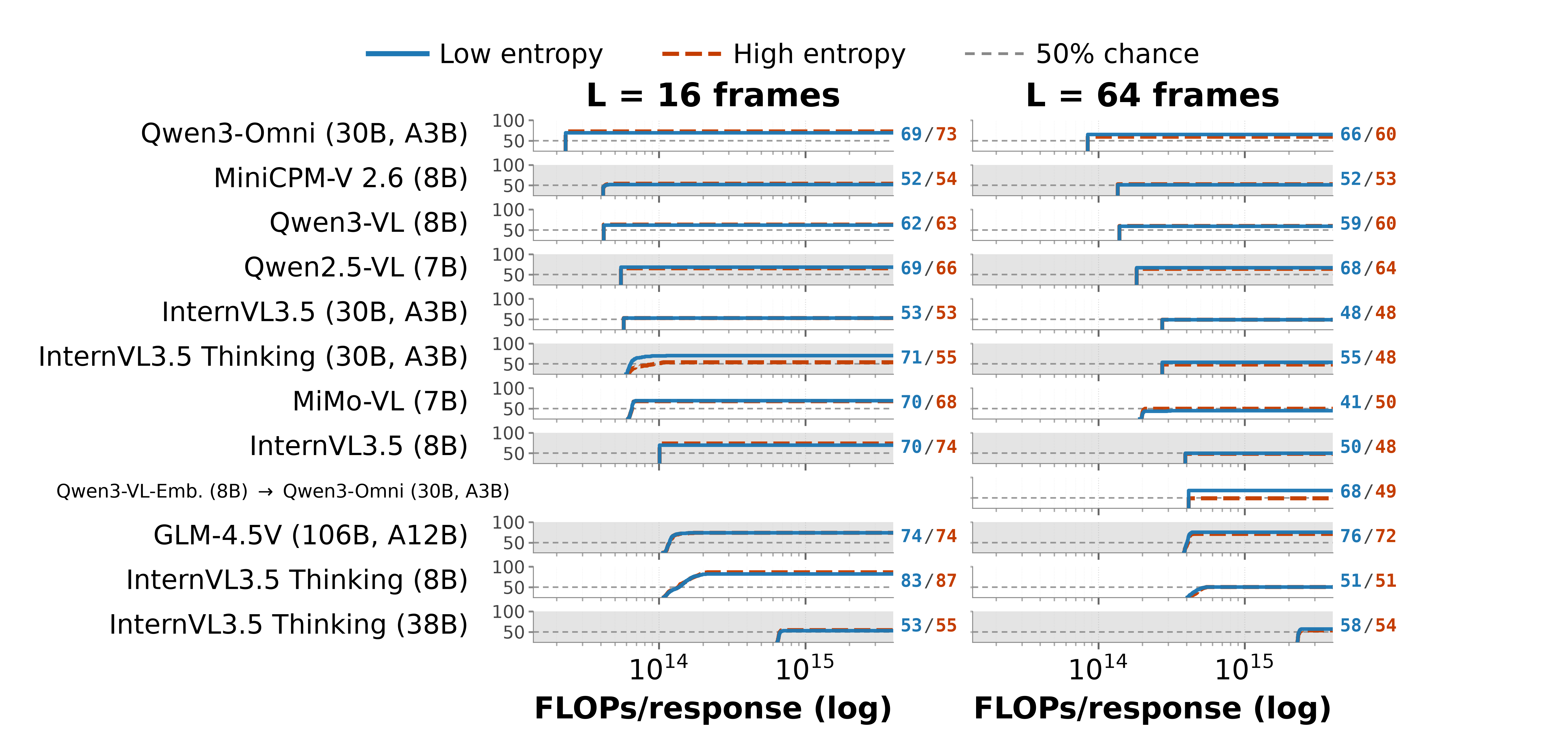}
    \caption{\textbf{VLMs cannot compress synthetic video.} Accuracy falls from 16 to 64 frames; at 64 frames only GLM-4.5V stays above 70\,\% accuracy, while most VLMs are inaccurate and do not score higher on predictable streams. Accuracy vs.\ FLOPs at $L{=}16$ (left) and $L{=}64$ (right) frames. Blue solid: low-entropy; orange dashed: high-entropy. Right-edge integers: final accuracy under unlimited FLOPs. FLOPs grow with the number of generated tokens and accuracy at a budget is the accuracy achieved if the model was cut off at that point. Models with low accuracy on a short number of frames are not plotted to ensure we only test models with strong perception on short clips (Appendix~\ref{app:short_L_perception}); the complete numbers are in Appendix~\ref{app:synth_numbers}.}
    \label{fig:flops-video-64-sequential}

\end{figure}

\subsubsection{Video}
\paragraph{VLMs do not compress their memory on synthetic video.} Our results for synthetic video are in Figure~\ref{fig:flops-video-64-sequential}. Most models perform only as accurately as chance at both entropy levels, and the models that are performant gain little from structure. Only one VLM, InternVL3.5 Thinking (30B, A3B), has a gap larger than ten points, although the retrieval-based system we tested (Appendix~\ref{app:rag}) compresses at $L{=}64$. Capacity is also, relative to length, much lower as compared to text.

\begin{figure}[!htb]
    \centering
    \includegraphics[width=0.71835\textwidth,trim={4bp 10.080bp 73.08bp 17.160bp},clip]{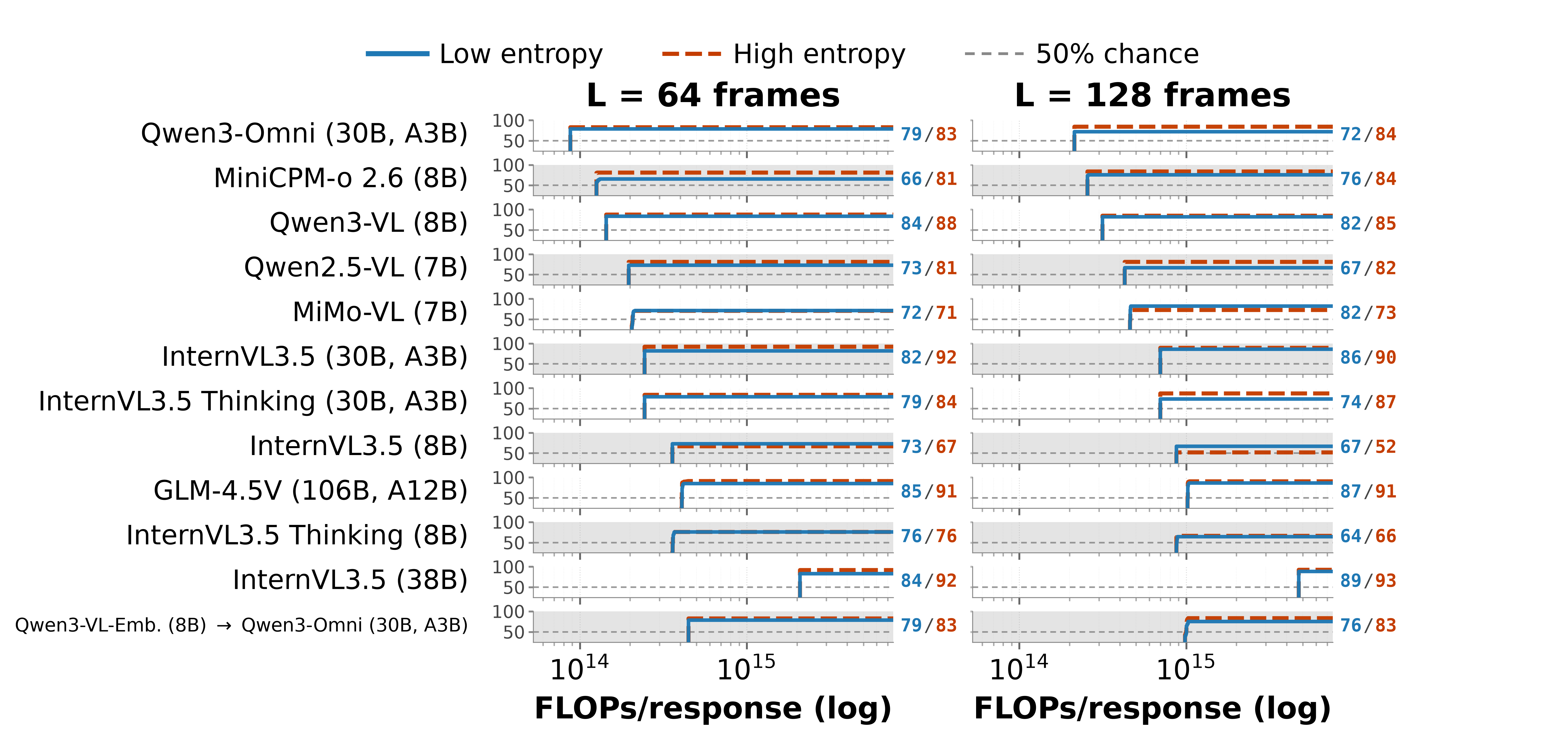}
    \caption{\textbf{On natural video, accuracy is higher than synthetic video, but gains from regularity are inconsistent.} Only one model has a large contrast; compare with Figure~\ref{fig:flops-1024-pretrained} where the contrast between high-entropy and low-entropy is much greater. We plot accuracy against FLOPs per response. Blue solid: low entropy; orange
    dashed: high entropy. FLOPs grow with the number of generated tokens and accuracy at a FLOPs budget is the accuracy achieved if the model was cut off at that point. The right-hand value is the cell's
    accuracy at maximum compute.}
    \label{fig:natvid_density}
\end{figure}

\paragraph{There is little evidence of natural-video compression.} Our results for natural video are in Figure~\ref{fig:natvid_density}. InternVL3.5 (8B) shows the largest low-entropy advantage at both lengths, gaining $5.9$ and $14.9$ points. However, it is the exception; in fact, many VLMs show anti-compression, or a higher accuracy on high-entropy videos than on those with low-entropy. The complete per-model contrasts are in Appendix Table~\ref{tab:band_contrast}.

\subsection{Are VLMs calibrated?}\label{sec:calibration}\label{sec:calibration_main}
Next, we ask if VLMs are calibrated and test this by repeating the prior experiments while explicitly specifying the cost of error~\citep{kalai2025hallucinate}. We define arm A, which has a payoff scheme of $+3$ for a correct answer and $-1$ for an incorrect answer, and arm B, which has a payoff of $+1$ and $-$4 for correct and incorrect answers respectively. The exact prompts used in these experiments are in Appendix~\ref{app:calibration_arms}.

\begin{figure}[!htb]
    \centering
    \includegraphics[width=0.8367\textwidth,trim={0 16bp 0 10bp},clip]{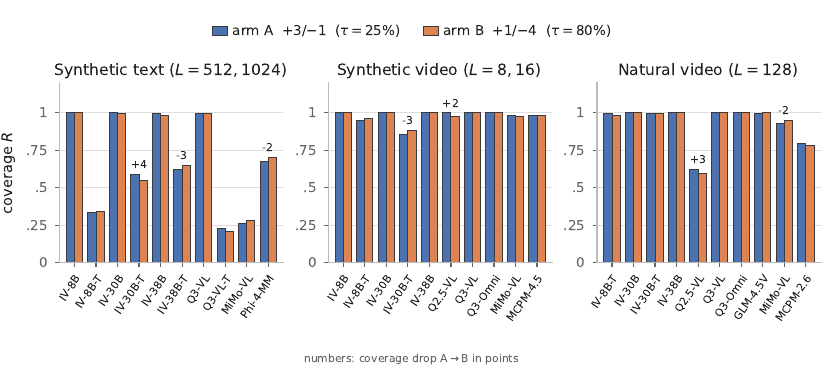}
    \caption{\textbf{VLMs are poorly calibrated on all modalities.} Coverage $R$ under each arm, per model, for all three subsets. No model abstains meaningfully more often as the payoff changes drastically. The numbers above bars show the coverage drop from arm A to arm B (shown when at least 2).}
    \label{fig:calib_coverage}
\end{figure}

\begin{figure}[!htb]
    \centering
    \includegraphics[width=0.927\textwidth,trim={0 5bp 0 5bp},clip]{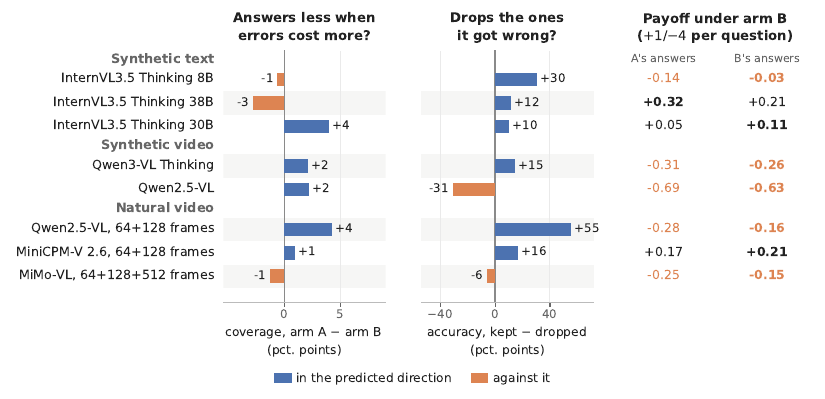}
    \caption{\textbf{VLMs only slightly optimize their payoffs by abstaining.} Each model is asked to answer given a payoff scheme of $+3$/$-1$ (arm A) or $+1$/$-4$ (arm B).
    \emph{Left}: does the model abstain more when a wrong answer costs four times
    as much? \emph{Middle}: was the model less accurate (on arm A) on the questions it dropped on arm B? \emph{Right}: average payoff per question with both arms scored under arm~B's payoff.}
    \label{fig:calib_resp_sel}
\end{figure}

\paragraph{VLMs respond weakly to increased cost of error, and score poorly.} The VLMs we tested do not significantly change their coverage when the penalty for incorrect answers increases (Figure~\ref{fig:calib_coverage}), and many models abstain more under the more favorable payoff scheme. Nevertheless, when both sets of responses are scored under arm B, most models obtain a higher payoff when prompted with the correct scoring scheme
  (Figure~\ref{fig:calib_resp_sel}, right). For most models we tested, questions they drop are generally ones they answered incorrectly under arm A (middle), indicating they are somewhat selective. However, this increase in payoff is small, and many models obtain negative scores, meaning they would be better off always abstaining.

\subsection{Do other backbones have better memory than RoPE Transformers?}\label{sec:backbone}
Our prior results suggest that VLMs lack several properties required for strong memory, especially on video, motivating us to test whether architecture is one source of this limitation. We therefore use a token-level version of ECCBench with the same task structure and vocabulary, training small models only on high-entropy streams of length $L \le 1024$ and evaluating their ability to generalize to longer, unseen streams ($L\in\{2048,4096\}$) at both high and low entropy. Although this is not a direct comparison with large pretrained VLMs, it isolates the memory properties of attention-based, linear-recurrent, state-space, memory-augmented, and recurrent backbones (Appendix Table~\ref{tab:backbone_configs}; Appendix~\ref{app:backbone_details}). Because these models cannot be told a payoff, we measure confidence using the output softmax score and analyze coverage--accuracy tradeoffs, with $L{=}2048$ shown in Figure~\ref{fig:backbone_calib_main} and $L{=}4096$ in Appendix Figure~\ref{fig:covacc_entropy_4096}.

\begin{figure}[!htb]
    \centering
    \includegraphics[width=0.922\textwidth,trim={0 54.000bp 0 55.000bp},clip]{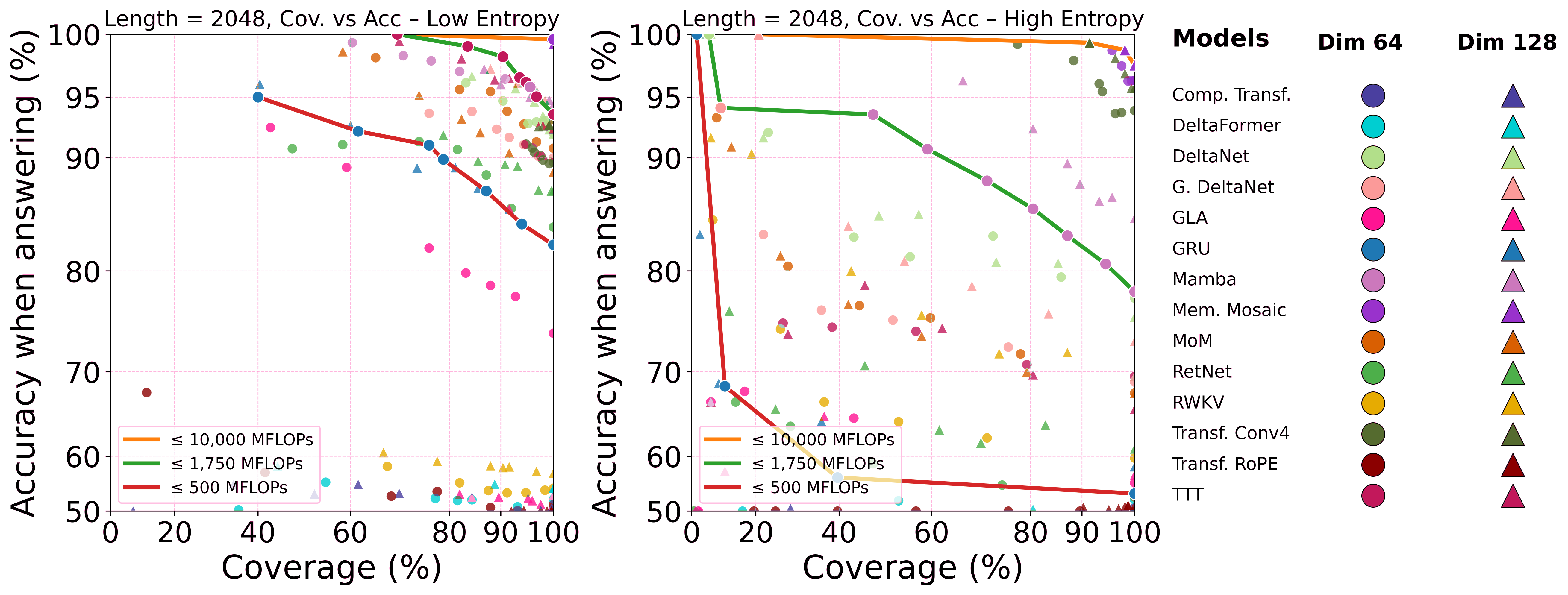}
    \caption{\textbf{Non-Transformers generally show better properties than RoPE Transformers.} Accuracy on answered queries versus coverage at ($L{=}2048$), after training on ($L\leq1024$). Left: low entropy; right: high entropy. Higher curves indicate more accurate recall at the same coverage; differences between panels show benefits from structured data. Circles and triangles denote model widths (D=64) and (128),
  respectively.}
    \label{fig:backbone_calib_main}
\end{figure}

\paragraph{Several models compress despite training only on incompressible noise.} Most surprising is that many models dynamically develop the ability to compress despite being trained only on incompressible noise. We can see this for individual models with high compressibility such as  RetNet and TTT-MLP (Figure~\ref{fig:backbone_calib_main}, $D=64$). We note that due to our definition of accuracy, near-perfect accuracy can make compression difficult to measure.

\paragraph{Non-Transformer backbones compress better than the RoPE Transformer.} We can read the two parts of Figure~\ref{fig:backbone_calib_main} together to compare accuracy--coverage frontiers on low and high entropy inputs. Notably, RoPE Transformers have extremely low capacity beyond training length, in line with the poor length extrapolation of rotary
encoding~\citep{kazemnejad2023nope}. Although replacing this positional encoding with a local convolution vastly increases capacity, their compression is outmatched by many models; DeltaNet, Test-Time Training Layers, and Mamba, among others, have pronouncedly higher accuracy on the left graph than on the right.

\paragraph{Non-Transformer models are better calibrated than the Transformer.} On both graphs in Figure~\ref{fig:backbone_calib_main}, RoPE Transformers lie near the bottom, and their coverage must be lowered drastically before accuracy increases above fifty percent. Although the Transformer with the convolution has a higher accuracy-coverage curve, it spans a narrow coverage range, so lowering coverage increases accuracy very little. In scenarios where incorrect answers are extremely expensive, it is sometimes desirable to abstain on nearly all queries except ones in which the memory contains a confident answer. Lowering coverage buys very little for the Transformer as compared to models like the GRU, TTT-MLP, or GLA, for instance.

\FloatBarrier
\section{Conclusion}
We introduced ECCBench, a framework for evaluating memory beyond accuracy. Our experiments reveal substantial weaknesses in pretrained VLMs, especially with compression, and demonstrate that the backbone architecture materially affects the properties of memory. 

As long-horizon agents become ubiquitous, memory will inevitably become a problem of managing limited resources. It is impossible to preserve all detail in an ever-growing stream, and general-purpose models cannot always predict which information will matter in the future. For example, current frontier models limit the maximum number of images they can ingest, so it is difficult to ascertain how well they would perform on much longer videos.\footnote{At the time of writing, the \href{https://ai.google.dev/gemini-api/docs/image-understanding}{Gemini}, \href{https://developers.openai.com/api/docs/guides/images-vision}{OpenAI}, and \href{https://platform.claude.com/docs/en/build-with-claude/vision}{Claude} APIs accept a maximum of 3,600, 1,500, and 600 images per request, respectively.} Even if they are highly accurate on short streams, if they cannot also compress, they will not be able to efficiently process data that spans months, years, or decades.

ECC is what makes processing such horizons possible, since a model that compresses well can degrade far less as a stream grows, and a calibrated model knows when its memory is insufficient and it must process more information.
We hope that our work inspires new approaches, architectures, and training procedures to create systems that possess ECC and perform well over ultra-long horizons.

%% file: appendix.tex
\tableofcontents
\vspace{1em}

\section{Benchmark Details}
\label{app:benchmark_details}

ECCBench is available at
\url{https://huggingface.co/datasets/anonstreammem/substream-recollection}. The Synthetic and EasyHuman content is
released under CC BY 4.0, and the EPIC-Kitchens-100-derived clips are provided
under CC BY-NC 4.0. For SoccerNet, we provide only provenance information, as the underlying videos are subject to an NDA and are available from the original authors. 

\subsection{Entropy-controlled generator}
\label{app:entropy_generator}

Streams are drawn over a fixed alphabet $\mathcal{V}$ with $|\mathcal{V}|=16$ symbols. Generation buckets use analytic entropy of variable-order $n$-gram rules (lengths 1--8). Backbone groups instead use empirical token entropy, where the low/high cutoffs are $2.0643/2.9451$ at $L{=}2048$ and $0.4904/3.5938$ at $L{=}4096$ (rounded). Low is at or below the first, high at or above the second.

In our synthetic experiments, we use two entropy bands that we use as examples of low- and high-entropy data:
\begin{itemize}
  \item \textbf{Low entropy:} streams with strong repetition. The text streams have mean analytic entropy $0.667$ bits/symbol (std $0.157$), range $[0.350,0.996]$.
  \item \textbf{High-entropy:} symbols sampled i.i.d.\ from the uniform distribution over the vocabulary; analytic entropy is $\log_2 16 = 4$ bits/symbol.
\end{itemize}

Here, a rule set is a small dictionary mapping contexts to distributions over the next item in the stream. Each context is a string of one to eight symbols from $\mathcal{V}$, meaning the generator is based on variable n-grams. For each context we specify up to four next-symbol probabilities, and the remaining probability mass spreads uniformly across the whole alphabet. For length $n>1$, $\hat h_{\mathrm{LZ}}=n\log_2 n/\sum_i\lambda_i$, where $\lambda_i$ is one plus the longest earlier match, allowing overlap and stopping at the stream end. To generate the next item in a stream, we find the longest suffix of the current prefix that matches some context in the rule set and then sample the next symbol from whatever the rule set defines as the output distribution. If there is no matching context, we draw uniformly over the vocabulary. To create the bands, we accept rule sets where the analytic entropy lies in the generation band.

Each (length, band) cell contains 20 sequences at all lengths that we evaluate. Each sequence is then paired with 16 questions which are a roughly equal mix of true positives and true negatives. Each candidate is three to six symbols long. A positive is a contiguous subsequence copied from a random position in the stream. A negative is sampled uniformly and is kept only if it does not occur anywhere in it. The video variant uses the same candidates rendered as frames.

Of course, repetition can help even a memory that stores random fragments, and learned rules can reject unlikely negatives without recalling exact occurrences. All models have these opportunities and may learn to exploit structure; we simply measure their benefit from regularity, because learning a rule is likely to help more than storing a recurring fragment.

Two examples of length 16, low entropy ($\hat h_{\mathrm{LZ}} \approx1.02$)
first and high entropy ($\log_2|\mathcal{V}| = 4.0$) second:
\begin{quote}
\texttt{8, 1, 4, 6, 2, 1, 1, 2, 1, 1, 2, 1, 1, 2, 1, 1}\\
\texttt{7, 1, 13, 13, 5, 2, 7, 13, 11, 9, 8, 7, 0, 0, 15, 12}
\end{quote}

\subsection{EasyHuman}\label{sec:easyhuman}
\begin{figure}[t]
\centering
\includegraphics[width=0.5\textwidth]{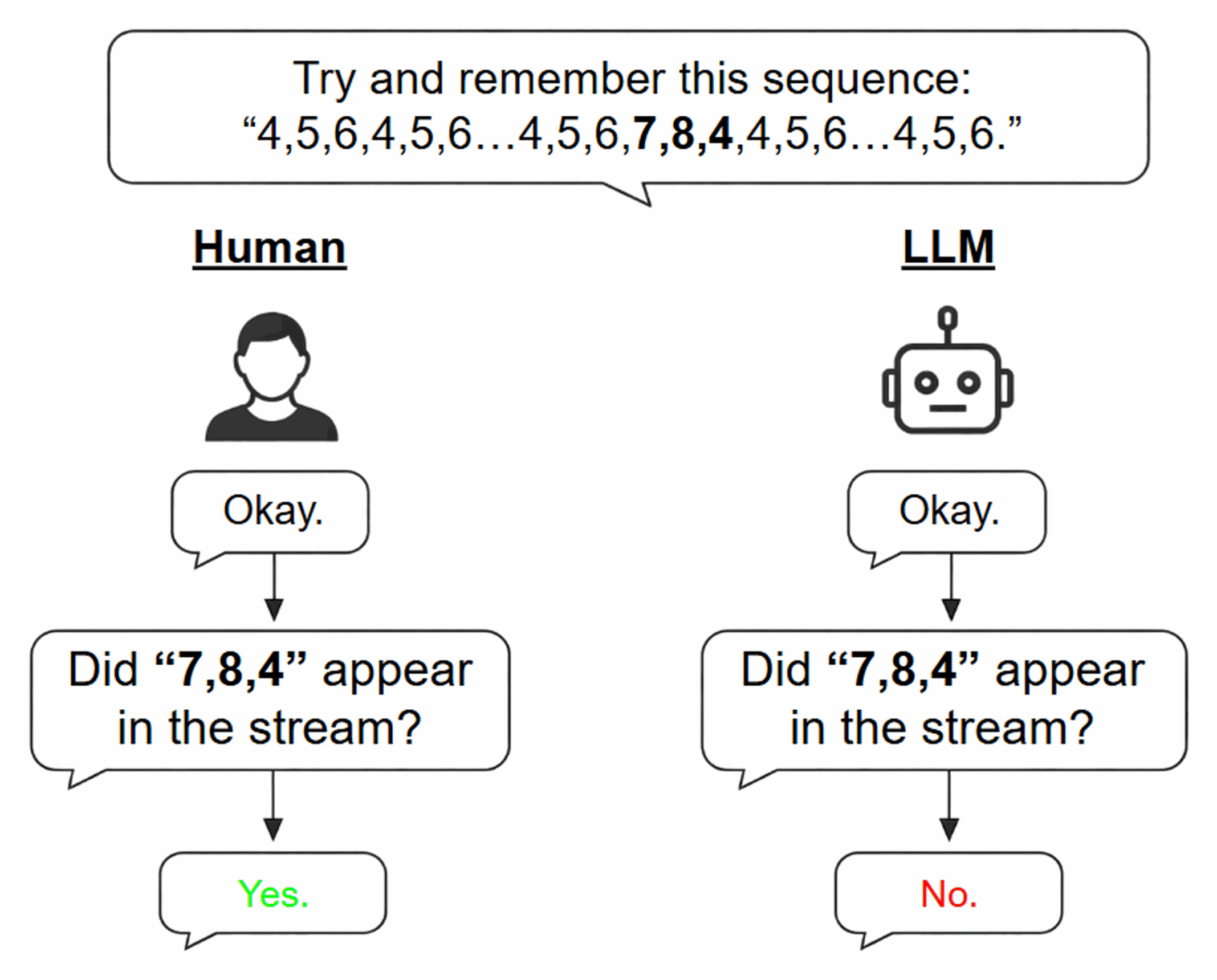}
\caption{\textbf{Models struggle as length grows despite repeating structure.} The EasyHuman task: a short pattern repeats with one deviation. No human study was run.}
\label{fig:easyhuman_example}
\end{figure}

One of the motivations of our task is the desire to have human-like memory, which has sensitivity to sequence violations~\citep{planton2021binarymemory} and better recall for some surprising events~\citep{foster2019surprise}. Each stream repeats a short pattern of length 3--5 with one or two short deviations, and queries target the deviation and corruptions of it (Figure~\ref{fig:easyhuman_example}; construction below). EasyHuman is named for how obvious its redundancy is to a person, but we did not run a human study.

\textbf{Models partly compress text and do not compress video at all.} Results are shown in Table~\ref{tab:easy_human_text_video_accuracy} for text and video, where $L$ denotes the stream length (in number of comma-separated numbers for text streams, in frames for video streams). Despite the extreme redundancy of the inputs, no open-weights model reliably exceeds 77\,\% accuracy in text, and video accuracy is near chance. For most models, text accuracy degrades only mildly from $L{=}256$ to $L{=}1024$, consistent with mild context rot. On video, models do very poorly and almost never abstain, which points to memory rather than perception for the models that pass the short-video check (Appendix~\ref{app:short_L_perception}).

\textbf{Frontier models also cannot compress video.} Claude Opus 4.7, Gemini 3.1 Pro and GPT-5.4 Pro all score above $90\,\%$ on EasyHuman text at both lengths, which is well above any open-weight model we tested. Still, they all perform poorly on video. We show their results in Table~\ref{tab:easy_human_text_video_accuracy} and because they are closed source and thus do not report FLOPs we do not use them in the compute-aware analysis.

\begin{table}[h]
  \centering
  \small
  \begin{tabular}{lccc}
  \toprule
  Model & Text $L{=}256$ & Text $L{=}1024$ & Video $L{=}256$ \\
  \midrule
  GLM-4.5V             & 76.3 & 70.5 & 49.1 \\
  InternVL3.5 (8B)     & 72.3 & 66.1 & 47.8 \\
  InternVL3.5 Thinking (30B, A3B) & 52.2 & 27.5 & 53.6 \\
  Phi-4-MM             & 62.5 & 59.4 & 56.2 \\
  Qwen3-VL             & 62.1 & 54.9 & 49.6 \\
  Qwen3-VL Thinking    & 11.6 &  9.8 & 53.1 \\
  MiMo-VL              & 41.1 & 37.1 & 52.7 \\
  \midrule
  \multicolumn{4}{l}{\emph{API-only; no FLOPs reported}} \\
  Claude Opus 4.7      & 91.6 & 90.9 & 51.1 \\
  Gemini 3.1 Pro       & 93.0 & 91.3 & 52.6 \\
  GPT-5.4 Pro         & 93.0 & 91.3 & 51.6 \\
  \bottomrule
  \end{tabular}
 \caption{\textbf{Models can compress text, but not videos.} EasyHuman accuracy (\%) for text at $L{=}256$ and $L{=}1024$ and video at $L{=}256$. Abstentions count as incorrect. No model's performance on video is distinguishable from chance, including the three closed-weights API models, which lead on text.}
  \label{tab:easy_human_text_video_accuracy}
\end{table}

\paragraph{Construction.}
\label{app:easyhuman_construction}
The EasyHuman subset consists of highly regular streams with a small number of deviations, and queries that focus on the deviation and its corruptions.

We use the following structures. In all cases, the deviations are chosen such that they do not already exist in the stream.
\begin{itemize}
    \item Repeating Pattern. A uniformly drawn motif of $3$--$5$ symbols is repeated to fill the stream, with one or two deviations of the same length placed anywhere in it.
    \item Counting Disruption. The base motif is a contiguous count by $1$ between two integers, either ascending (e.g., $2,3,4,5$) or descending (e.g., $13,12,11$); motif lengths are $3$--$5$ symbols, and the motif is repeated to fill the stream. Deviations are also $3$--$5$ symbols long and are drawn from the uniform distribution. In video, random symbol-to-letter mapping preserves repetition but need not preserve alphabetical order.
\end{itemize}

\subsection{Video rendering and question variants}
\label{app:video_rendering}
Each symbol becomes one letter on one of three vertical conveyor belts (Figure~\ref{fig:data-examples}b), rendered at $448\times448$ pixels and 1 frame per second, so a length-$L$ stream is an $L$-frame, $L$-second video. The belt positions of each symbol come from a separately generated three-symbol sequence of a similar entropy level; it is structured in the low-entropy split and sampled uniformly in the high-entropy split. A letter can therefore appear on different belts within one stream.

\paragraph{Why glyphs and their
  arrangement must be retained but not other details}
The queries concern letter order and belt identity, not other visual details such as belt texture. Each symbol occurrence occupies one frame, in the main video and every candidate alike, with no padding or extra frames. All candidates use the same rendering style.

\paragraph{Glyph choice in text and video}
We use numbers (0-15) for text modality and letters for the video modality. We choose this because in preliminary trials that we ran, we found that VLMs read rendered letters off video frames more reliably than rendered digits at the same font size. Keeping the underlying alphabet at 16 symbols, we picked the most readable rendering for each modality, and do not believe that changing these decisions would significantly impact our results.

The synthetic video subset has two question variants which are pooled in our main results. Both have candidates which are separate videos rendered using the same background, fps, and other parameters as the main video (except they are much shorter). Separate results are in Table~\ref{tab:video_question_types}:
\begin{itemize}
    \item \textbf{Sequential.} A true positive sequential candidate is a clip taken from the original video. A false candidate may have different glyphs appear, but will never show a sequence of glyphs which did appear in the original video scrambled onto different belt positions.
    \item \textbf{Spatial.} A true positive, like in sequential, is in the original sequence. A negative preserves the letters of a sampled subsequence but resamples the glyphs' belt positions.
\end{itemize}

\begin{table}[!htb]
\centering
\small
\begin{tabular}{l rr rr rr rr}
\toprule
& \multicolumn{4}{c}{$L{=}16$} & \multicolumn{4}{c}{$L{=}64$} \\
\cmidrule(lr){2-5}\cmidrule(lr){6-9}
& \multicolumn{2}{c}{low entropy} & \multicolumn{2}{c}{high entropy} & \multicolumn{2}{c}{low entropy} & \multicolumn{2}{c}{high entropy} \\
\cmidrule(lr){2-3}\cmidrule(lr){4-5}\cmidrule(lr){6-7}\cmidrule(lr){8-9}
Model & seq. & spatial & seq. & spatial & seq. & spatial & seq. & spatial \\
\midrule
GLM-4.5V (106B, A12B) & 73.6 & 73.9 & 74.1 & 73.5 & 73.9 & 77.3 & 78.0 & 65.6 \\
InternVL3.5 (30B, A3B) & 54.1 & 52.8 & 52.5 & 53.7 & 47.1 & 49.7 & 46.6 & 50.3 \\
InternVL3.5 (8B) & 84.9 & 54.7 & 88.0 & 60.5 & 52.2 & 47.9 & 46.6 & 50.3 \\
InternVL3.5 Thinking (30B, A3B) & 72.3 & 68.9 & 52.5 & 56.8 & 60.9 & 48.5 & 54.7 & 41.5 \\
InternVL3.5 Thinking (38B) & 52.2 & 53.4 & 58.2 & 51.2 & 59.9 & 55.2 & 55.3 & 52.2 \\
InternVL3.5 Thinking (8B) & 85.5 & 80.1 & 90.5 & 83.3 & 51.6 & 49.7 & 53.4 & 48.4 \\
LongVILA (7B) & 46.5 & 44.7 & 49.4 & 42.0 & 39.5 & 38.0 & 27.3 & 27.0 \\
MiMo-VL (7B) & 71.1 & 68.3 & 74.7 & 62.3 & 37.6 & 43.6 & 47.2 & 53.5 \\
MiniCPM-V 2.6 (8B) & 54.1 & 50.9 & 63.3 & 45.1 & 55.4 & 47.9 & 53.4 & 52.8 \\
MiniCPM-V 4.5 (9B) & 62.9 & 55.3 & 59.5 & 51.2 & 63.7 & 47.2 & 50.9 & 48.4 \\
Phi-4-MM (6B) & 48.4 & 41.0 & 43.7 & 36.4 & 43.9 & 49.7 & 34.8 & 39.6 \\
Qwen2.5-VL (7B) & 67.9 & 69.6 & 64.6 & 67.3 & 70.7 & 64.4 & 66.5 & 62.3 \\
Qwen3-Omni (30B, A3B) & 76.7 & 62.1 & 80.4 & 65.4 & 75.8 & 55.8 & 64.6 & 55.3 \\
Qwen3-VL (8B) & 64.2 & 59.6 & 63.9 & 63.0 & 60.5 & 58.3 & 60.2 & 59.7 \\
Qwen3-VL Thinking (8B) & 55.3 & 43.5 & 46.2 & 43.2 & 47.8 & 34.4 & 46.6 & 41.5 \\
\bottomrule
\end{tabular}
\caption{Synthetic-video accuracy (\%) split into sequential and spatial questions. The pooled scores weight each question equally, but the two variants have slightly different counts.}
\label{tab:video_question_types}
\end{table}

The model receives the same prompt format in both variants. The distinction lies only in how candidates are constructed and the types of incorrect answers. For both types of questions, a match must preserve symbol order and belt identity for a candidate to be a positive match.

\subsection{Dataset documentation}
\label{app:dataset_documentation}
A Croissant 1.0 \texttt{metadata.json} in the HuggingFace release is the datasheet for ECCBench and covers the standard Datasheets-for-Datasets categories~\citep{gebru2021datasheets}, including the Responsible-AI fields. The release also contains the manifest that is needed for our codebase: a top-level \texttt{questions\_dataset.json} index and a per-bucket \texttt{manifest.json}. Errata and reproduction issues may be reported through the issue tracker of the code repository at \url{\codeurl}, linked from the dataset card. Corrections and expanded splits will appear in the HuggingFace revision history.

\FloatBarrier
\section{Pretrained VLMs Experimental Details}
\label{app:pretrained_eval_details}

\subsection{Evaluation prompts}
\label{app:prompts}
We use the following three prompts. The model can answer \texttt{\{0\}} for yes, \texttt{\{1\}} for no, or \texttt{\{2\}} for abstain. We parse the outputs by extracting the last bracketed token in its output text, and if no bracketed answer is emitted, the response is treated as an implicit abstention.\footnote{The video prompt below uses the spelling ``conveyer'' rather than the standard ``conveyor''; this misspelling was present in the prompt seen by every model.}

\paragraph{Prompt for video questions.}
\begin{verbatim}
"In the main video and option video, letters moved down three
 conveyer belts."
"A letter that appeared in any of the videos on a specific
 conveyer belt ALWAYS stayed on the same conveyer belt and
 only moved down, not left or right."
"The letters appeared sequentially. Only one was on the screen
 at a time."
"The letters continued moving down the conveyer belt until
 they left the screen."
"Examine the main video and the option video closely."
"Did this option video show events that exactly appeared
 SOMEWHERE in the main video, down to the order they appeared
 and which conveyer belt each letter was on?"
"Write {0} if yes, {1} if no, or {2} if unsure.
 Please put your answer in curly braces, e.g, {2}."
\end{verbatim}

\paragraph{Prompt for text questions.}
\begin{verbatim}
"You are given a main sequence of tokens and a candidate
 sequence."
"Your task: Determine if the candidate sequence appears as a
 contiguous subsequence within the main sequence, in the exact
 same order."
"The candidate must appear with all its tokens consecutive and
 in order - no gaps, no extra tokens in between."
"Answer {0} if the candidate appears as a contiguous
 subsequence, {1} if it does not, or {2} if you are unsure.
 Please put your answer in curly braces, e.g, {2}."
\end{verbatim}

\paragraph{Prompt for natural-video questions.} 
\begin{verbatim}
Watch this video closely, paying attention to the entire video.
{question}

Write {0} if yes, {1} if no or {uncertain} if you are unsure.
Please put your answer in curly braces, e.g, {0}.
\end{verbatim}

\subsection{Compute resources}
\label{app:compute_resources}
All experiments ran on an internal cluster of A40, H200, L40, and RTX 3090 GPUs. The pretrained VLM evaluation jobs each allocated 2--7 GPUs per job depending on the model they evaluated. Each backbone job used a 36 wall-clock-hour training budget across four runs per architecture, on single-GPU nodes (Section~\ref{app:backbone_details}). The full project consumed approximately 60{,}000 GPU-hours.

\subsection{Per-model FLOPs equations}
\label{app:flops_methodology}
\label{app:flops_per_model}

We count the FLOPs analytically. For
each response we evaluate a formula whose inputs are the tokens in the query and the model's architectural constants
(Table~\ref{tab:flops_constants}). Our aim in calculating FLOPs like so is a like-for-like
comparison of how much arithmetic each architecture must perform to access its memory.

\paragraph{Constants.} We obtain architectural constants from each model's checkpoint configuration on HuggingFace or its implementation if it was released elsewhere.

\begin{table}[tbp]
\centering
\small
\begin{tabular}{lrrrrrrr}
\toprule
Model & $L_v$ & $d_v$ & $f_v$ & $L_t$ & $d_t$ & $n_q/n_{kv}$ & $f_t$ \\
\midrule
Qwen2.5-VL (7B)                  & 32 & 1280 & 3420 & 28 & 3584 & 28/4 & 18944 \\
Qwen3-VL (8B) (+Thinking)        & 27 & 1152 & 4304 & 36 & 4096 & 32/8 & 12288 \\
Qwen3-Omni (30B, A3B)            & 27 & 1152 & 4304 & 48 & 2048 & 32/4 & 768 \\
GLM-4.5V (106B, A12B)            & 24 & 1536 & 4096 & 46 & 4096 & 96/8 & 1408 \\
InternVL3.5 (8B) (+Thinking)     & 24 & 1024 & 4096 & 36 & 4096 & 32/8 & 12288 \\
InternVL3.5 (30B, A3B) (+Think.) & 24 & 1024 & 4096 & 48 & 2048 & 32/4 & 768 \\
InternVL3.5 (38B) (+Thinking)    & 45 & 3200 & 12800 & 64 & 5120 & 64/8 & 25600 \\
MiMo-VL (7B)                     & 32 & 1280 & 3456 & 36 & 4096 & 32/8 & 11008 \\
MiniCPM-V 4.5 (9B)               & 27 & 1152 & 4304 & 36 & 4096 & 32/8 & 12288 \\
MiniCPM-V 2.6 (8B)               & 27 & 1152 & 4304 & 28 & 3584 & 28/4 & 18944 \\
LongVILA (7B)                    & 27 & 1152 & 4304 & 28 & 3584 & 28/4 & 18944 \\
Phi-4-MM (6B)                    & 27 & 1152 & 4304 & 32 & 3072 & 24/8 & 8192 \\
\bottomrule
\end{tabular}

\medskip

\begin{tabular}{lrrl}
\toprule
Model & MoE & $V_{\mathrm{voc}}$ & FFN style \\
\midrule
Qwen2.5-VL (7B)                  & -- & 152064 & SG / SG \\
Qwen3-VL (8B) (+Thinking)        & -- & 151936 & SG / G \\
Qwen3-Omni (30B, A3B)            & 128/8 & 152064 & MoE SG / G \\
GLM-4.5V (106B, A12B)            & 128/8+sh & 151552 & MoE SG / SG \\
InternVL3.5 (8B) (+Thinking)     & -- & 151936 & SG / G \\
InternVL3.5 (30B, A3B) (+Think.) & 128/8 & 151936 & MoE SG / G \\
InternVL3.5 (38B) (+Thinking)    & -- & 151936 & SG / G \\
MiMo-VL (7B)                     & -- & 151680 & SG / SG \\
MiniCPM-V 4.5 (9B)               & -- & 151748 & SG / G \\
MiniCPM-V 2.6 (8B)               & -- & 151666 & SG / G \\
LongVILA (7B)                    & -- & 151651 & SG / G \\
Phi-4-MM (6B)                    & -- & 200064 & SG / G, + LoRA $r{=}256$ \\
\bottomrule
\end{tabular}
\caption{Constants entering the per-response FLOPs formulas, read from each
model's released checkpoint configuration. $L_v, d_v, f_v$: vision
transformer depth, width, feed-forward width. $L_t, d_t, n_q/n_{kv},
f_t$: language model depth, width, query / key-value head counts, and
feed-forward width (per expert where the model is
mixture-of-experts). ``MoE'' gives the expert count $E$ and the number active
per token, with ``+sh'' marking an additional always-active shared expert.
$V_{\mathrm{voc}}$ is the vocabulary size, which is used only through the
output projection $F_{\mathrm{voc}} = 2 d_t V_{\mathrm{voc}}$. ``FFN style'' gives the
language model's / the vision tower's feed-forward form: G = two matrices,
SG = three matrices (gated). }
\label{tab:flops_constants}
\end{table}

\paragraph{FLOPs calculations.} A matrix multiplication of shapes $(A,B) \cdot (B,C)$
costs $2ABC$ FLOPs, two operations per multiply--add, and that is all we
count. We do not count elementwise operations such as normalization, softmax, activation
functions, rotary position rotations, residual additions, or embedding lookups.
\paragraph{Cost terms.} Although there are standard, simpler ways of counting per-model FLOPs in LLMs, the vision encoder makes these more complicated. Additionally, the small differences in how these models chunk up data can make enormous differences on video data. We thus decide to use analytical formulas for FLOPs computation which take these into account. We can split the cost of a model into the vision encoder,
the connector that maps vision features into the language model's input
space, the language model's reading of the prompt (prefill), and its
generation of the answer (decode):
\[
F_{\mathrm{total}}(L, T_{\mathrm{in}}, T_{\mathrm{out}})
\;=\;
F_{\mathrm{vis}} \;+\; F_{\mathrm{conn}} \;+\; F_{\mathrm{pre}} \;+\; F_{\mathrm{dec}},
\]
for $L$ input frames, $T_{\mathrm{in}}$ prompt tokens and $T_{\mathrm{out}}$
generated tokens. Let $d_v$, $f_v$, and $L_v$ denote the vision transformer's hidden dimension, FFN width, and number of blocks, respectively, with $V$ representing the sequence length per vision block. 
For the language model, $d_t$ and $L_t$ denote its hidden dimension and layer count, $n_q$ and $n_{kv}$ its query and key/value head counts, and $h_t$ its head dimension (fixed at $h_t = 128$ across all the models we evaluated), which gives query and key/value projections of dimensions $d_q = n_q h_t$ and $d_{kv} = n_{kv} h_t$. 
For mixture-of-experts (MoE) architectures, $f_t$ denotes the per-expert FFN width, $E$ the total number of experts, and $k$ the number of active experts per token. 
$V_{\mathrm{voc}}$ is the vocabulary size, $\Phi$ represents the per-token FFN cost, and $N = T_{\mathrm{in}} + V_{\mathrm{LLM}}$ is the total sequence length processed by the language model, where $V_{\mathrm{LLM}}$ is the number of vision tokens produced by the visual connector.

\paragraph{Vision encoder.} Each frame is cut into square patches with each
patch becoming a token. We set a vision block processing $V$ tokens, each
attending over $S$ positions, as costing
\[
F_{\mathrm{blk}}(V, S)
\;=\;
\underbrace{8 V d_v^2}_{\text{query, key, value, output projections}}
\;+\;
\underbrace{4 V S d_v}_{\text{attention itself}}
\;+\;
F_{\mathrm{FFN}}(V),
\]
\[
F_{\mathrm{FFN}}(V) =
\begin{cases}
4 V d_v f_v & \text{two-matrix feed-forward},\\
6 V d_v f_v & \text{three-matrix (gated) feed-forward},
\end{cases}
\]
and the tower also has to pay the cost of turning raw pixels into patch features,
$2 V_{\mathrm{pre}} d_{\mathrm{pat}} d_v$. Here $V_{\mathrm{pre}}$ is the
number of patches before the class token is prepended, and
$d_{\mathrm{pat}}$ is the number of pixel values per patch: $3 p_v^2$ for
patch size $p_v$, doubled where two frames are packed together.

The token counts below describe synthetic frames at $448\times448$. For our natural videos which have a resolution of  $640\times360$, Qwen, MiMo, and GLM round each side to the nearest multiple of twice the patch size before $2\times2$ merging. This gives $V=1196$ patches and 299 language tokens per two-frame group for Qwen2.5, MiMo, and GLM, or $V=880$ and 220 for Qwen3. We use these numbers for the costs given in Table~\ref{tab:prefill_decode}.

\begin{table}[!htb]
\centering
\small
\begin{tabular}{l rrr rrr rrr}
\toprule
& \multicolumn{3}{c}{synthetic text} & \multicolumn{3}{c}{synthetic video, $448^2$} & \multicolumn{3}{c}{natural video, $640{\times}360$} \\
\cmidrule(lr){2-4}\cmidrule(lr){5-7}\cmidrule(lr){8-10}
Model & $N$ & $F_{\mathrm{pre}}$ & $F_{\mathrm{dec}}$ & $N$ & $F_{\mathrm{pre}}$ & $F_{\mathrm{dec}}$ & $N$ & $F_{\mathrm{pre}}$ & $F_{\mathrm{dec}}$ \\
 & tok & TF & GF & tok & TF & GF & tok & TF & GF \\
\midrule
Qwen2.5-VL (7B) & $64$ & $0.84$ & $29$ & $9{,}088$ & $135$ & $37$ & $9{,}696$ & $145$ & $37$ \\
Qwen3-VL (8B) & $64$ & $0.89$ & $32$ & $6{,}988$ & $111$ & $40$ & $7{,}168$ & $115$ & $40$ \\
Qwen3-Omni (30B, A3B) & $64$ & $0.35$ & $13$ & $6{,}988$ & $57$ & $24$ & $7{,}168$ & $59$ & $24$ \\
GLM-4.5V (106B, A12B) & $64$ & $1.56$ & $53$ & $9{,}088$ & $315$ & $94$ & $9{,}696$ & $343$ & $96$ \\
InternVL3.5 (8B) & $64$ & $0.89$ & $32$ & $17{,}792$ & $341$ & $53$ & $16{,}512$ & $310$ & $51$ \\
InternVL3.5 (30B, A3B) & $64$ & $0.35$ & $13$ & $17{,}792$ & $222$ & $41$ & $16{,}512$ & $197$ & $39$ \\
InternVL3.5 (38B) & $64$ & $4.00$ & $130$ & $17{,}792$ & $1{,}442$ & $204$ & $16{,}512$ & $1{,}316$ & $199$ \\
MiMo-VL (7B) & $64$ & $0.82$ & $29$ & $9{,}088$ & $140$ & $40$ & $9{,}696$ & $151$ & $41$ \\
MiniCPM-V 4.5 (9B) & $64$ & $0.89$ & $32$ & $4{,}544$ & $69$ & $37$ & $4{,}224$ & $64$ & $36$ \\
MiniCPM-V 2.6 (8B) & $64$ & $0.84$ & $29$ & $4{,}544$ & $63$ & $33$ & $4{,}224$ & $59$ & $33$ \\
LongVILA (7B) & $64$ & $0.84$ & $29$ & $4{,}497$ & $63$ & $33$ & $4{,}240$ & $59$ & $33$ \\
Phi-4-MM (6B) & $64$ & $0.41$ & $17$ & $37{,}733$ & $551$ & $48$ & $35{,}008$ & $492$ & $46$ \\
\bottomrule
\end{tabular}
\caption{What reading and writing cost, for every VLM with $64$ input frames (or text symbols) and a three-token answer. $N$ is the input length in tokens.}
\label{tab:prefill_decode}
\end{table}

$S$ represents how the cost of vision grows as the video gets longer:
\begin{itemize}
\item \textbf{Qwen2.5-VL, MiMo-VL.} Two consecutive frames are packed into one
temporal group of $V = 1024$ tokens. Four layers attend to the whole group, but the remaining 28 attend only within 64-token windows ($S = 64$).
\item \textbf{Qwen3-VL, Qwen3-VL Thinking, Qwen3-Omni.} All 27 blocks attend
within one two-frame temporal group of $V = S = 784$ tokens.
\item \textbf{GLM-4.5V.} $V = S = 1024$, one two-frame temporal group.
\item \textbf{InternVL3.5.} $V = S = 1025$ per $448^2$ tile: 1024 patches plus
one class token.
\item \textbf{MiniCPM-V 2.6 / 4.5, LongVILA, Phi-4-MM.} Each pass through the vision encoder processes a single image crop of size $V = S = 1024$ tokens. Phi-4-MM, however, explicitly divides each video frame into two crops.
\end{itemize}
Because each span is fixed, the number of spans grows proportionally with $L$, the frame count. 

\paragraph{Tokens inputted to the language model.} $V_{\mathrm{LLM}}$
follows each model's own preprocessing:
\begin{itemize}
\item \textbf{Qwen2.5-VL, MiMo-VL, GLM-4.5V.} Video frames are organized into temporal groups of two frames. Within each group, adjacent $2 \times 2$ patch grids are merged to one token producing $256$ tokens per temporal group and $256 \lceil L/2 \rceil$ visual tokens in total across all $L$ frames.
\item \textbf{Qwen3-VL, Qwen3-Omni.} Two frames per group, $2 \times 2$
merging over a $28 \times 28$ patch grid: $196 \lceil L/2 \rceil$.
\item \textbf{InternVL3.5.} Each video frame is processed as a single $448 \times 448$ tile (spatially). A pixel-unshuffle operation then downsamples each $2 \times 2$ grid of patch features into a single token, compressing 1,024 initial patches down to 256 tokens per frame, for a total sequence length of $256 L$.
\item \textbf{MiniCPM-V 2.6.} A cross-attention resampler uses 64 pre-learned query vectors to condense each frame's patch features into 64 tokens. This yields $64 L$ visual tokens.

\item \textbf{MiniCPM-V 4.5.} A spatio-temporal resampler processes frames in packs of size $p = \min\!\left(6, \max(1, \lceil L/180 \rceil)\right)$. Afterwards, each pack is compressed into 64 tokens, resulting in $V_{\mathrm{LLM}} = 64 \lceil L/p \rceil$ total tokens. 

\item \textbf{LongVILA.} Each frame is initially mapped to 256 tokens, then temporally averaged into groups of size $\mathrm{pool}(L) = 4 \cdot \mathrm{bucket}(L/256)$, where $\mathrm{bucket}(\cdot)$ rounds up to the nearest power of two in $\{1, 2, 4, 8\}$. Adding one separator token per pooled group gives $V_{\mathrm{LLM}} = 257 \lfloor L / \mathrm{pool}(L) \rfloor$ total tokens.

\item \textbf{Phi-4-MM.} Each frame is processed via one global view and one local crop view. Spatial pooling and inserted row-separator tokens give a cost of 545 tokens per frame ($545 L$ total).
\end{itemize}

\paragraph{Connector FLOPs.} 
We calculate the per-token FLOPs used by each model's connector between the vision and language towers. Let $m = 4 d_v$ represent the feature dimension of a merged $2 \times 2$ patch block:
\begin{itemize}[leftmargin=*, noitemsep]
    \item \textbf{Qwen2.5-VL, MiMo-VL, LongVILA:} They use a patch-merging layer costing $2 m^2 + 2 m d_t$ per token.
    \item \textbf{GLM-4.5V:} Follows its initial projection with a gated FFN of width $10944$, yielding a total cost of $2 m d_t + 2 d_t^2 + 6 d_t \cdot 10944$.
    \item \textbf{Qwen3:} Uses a primary merger alongside three DeepStack adapters---which inject vision-tower features into the decoder layers---which in total cost four full merger passes at $4 \times (2 m^2 + 2 m d_t)$.
    \item \textbf{InternVL3.5:} After pixel-unshuffling, it uses a two-layer linear projection costing $2 m d_t + 2 d_t^2$.
    \item \textbf{Phi-4-MM:} Performs spatial projection and pooling at a cost of $2 d_v d_t + 2 d_t^2$.
    \item \textbf{MiniCPM-V (2.6 / 4.5):} Uses a cross-attention resampler. For $K$ patch tokens mapped to 64 output tokens, the cross-attention cost is $2 K d_v d_t + 4 K d_t^2 + 256 d_t K + 384 d_t^2$.
\end{itemize}

\paragraph{The cost of prefill.} 
We decompose the prefill FLOP cost into feed-forward/projection overhead $B$ and causal attention $A$ for each layer:
\[
B \;=\; L_t \Bigl[\, 2 d_t (d_q + 2 d_{kv}) \;+\; 2 d_q d_t \;+\; \Phi \Bigr],
\qquad
A \;=\; 4 L_t n_q h_t .
\]
Processing a context of $N$ tokens then requires $F_{\mathrm{pre}}(N)$ FLOPs:
\[
F_{\mathrm{pre}}(N) \;=\; B\,N \;+\; A\,P(N),
\]

\paragraph{Generating the answer (decode).} Generation reuses the KV cache. Reading the prompt already produces the
distribution for the first generated token, so producing $T_{\mathrm{out}}$
tokens applies the output vocabulary projection $T_{\mathrm{out}}$ times but
runs the layers only $T_{\mathrm{out}} - 1$ times, the $t$-th of those
attending over $N + t$ cached positions:
\[
\begin{aligned}
F_{\mathrm{dec}}(N, T_{\mathrm{out}})
&= T_{\mathrm{out}} F_{\mathrm{voc}} + (T_{\mathrm{out}} - 1) B \\
&\quad + A\Bigl[(T_{\mathrm{out}} - 1) N + \tfrac{1}{2} T_{\mathrm{out}} (T_{\mathrm{out}} - 1)\Bigr], \\
F_{\mathrm{voc}} &= 2 d_t V_{\mathrm{voc}}.
\end{aligned}
\]
with the vocabulary projection $F_{\mathrm{voc}}$ we count as part of decode.

\paragraph{Per-token feed-forward cost.} For models with one gated
feed-forward per layer, $\Phi = 6 d_t f_t$. For mixture-of-experts models,
where a small router scores the experts and only the top $k$ run,
\[
\Phi \;=\; \underbrace{2 d_t E}_{\text{router}} \;+\; k \cdot 6 d_t f_t
\;+\; \mathbb{1}_{\mathrm{shared}} \cdot 6 d_t f_s ,
\]
the last term covering an always-active shared expert where the checkpoint has
one (GLM-4.5V, $f_s = 1408$; Qwen3-Omni and InternVL3.5 (30B, A3B) have
none). Two models are not uniform down their depth. GLM-4.5V's first decoder
layer is an ordinary feed-forward of width $10944$ and its remaining $45$
layers are mixture-of-experts, so its body cost is
$L_t \bigl[2 d_t (d_q + 2 d_{kv}) + 2 d_q d_t\bigr] + 6 d_t \cdot 10944 +
(L_t - 1)\Phi$. Phi-4-MM's vision LoRA adapters ($r = 256$) are dynamically used at inference and are then added to the cost of every layer.

For synthetic video, we estimate the cost of the candidate clip using five frames, although our candidate clips vary in length. Its vision and connector costs are added separately, and its tokens are included in the context read by the language model.

\paragraph{The cost of a response.} The prompts request a bracketed answer, such as \texttt{\{0\}} for yes (Section~\ref{app:prompts}). We count output through the last complete answer marker, including the marker itself, and use this length, $T_{\mathrm{ans}}$, in the decoding-cost formula above.

\paragraph{Reading and writing costs.} Recall in Section~\ref{sec:background} we defined $F = F^{\mathrm w} + F^{\mathrm r}$ as the total cost of memory. Our accounting assumes one query per stream. For a VLM, which takes the whole
stream into its context when the question arrives, the write cost of
Section~\ref{sec:background} includes the vision encoder, connector, and prefill of Section~\ref{app:flops_methodology}
and the read cost is the decode.

The cost of the reading operation can be divided into two parts. We write the per-token FLOPs cost, which is independent of sequence length, as $B$. Attention, though, grows quadratically, and the number
of pairs the model has to score is $P(N) = \tfrac{1}{2}N(N+1)$. We write the FLOPs cost of a pair as $A$. Reading
therefore costs
\[
F_{\mathrm{pre}}(N) \;=\; \underbrace{B\,N}_{\text{grows with }N}
\;+\; \underbrace{A\,P(N)}_{\text{grows with }N^{2}}.
\]

In general, reading dominates, especially on video. While there are some exceptions, notably reasoning models, many more tokens must be processed during the input than during the decode phase. Table~\ref{tab:flops_coeffs} shows the three FLOPs cost variables for each VLM and Table~\ref{tab:prefill_decode} shows the writing and answering costs. Units are given in millions (MF), billions (GF), and trillions (TF) of operations.

\begin{table}[!htb]
\centering
\small
\begin{tabular}{lrrr}
\toprule
Model & $B$ (MF/token) & $A$ (kF/pair) & $F_{\mathrm{voc}}$ (MF) \\
\midrule
Qwen2.5-VL (7B)~\citep{bai2025qwen25vl} & $13{,}051$ & $401$ & $1{,}090$ \\
Qwen3-VL (8B)~\citep{bai2025qwen3vl} & $13{,}892$ & $590$ & $1{,}245$ \\
Qwen3-Omni (30B, A3B)~\citep{xu2025qwen3omni} & $5{,}461$ & $786$ & $623$ \\
GLM-4.5V (106B, A12B)~\citep{glmv2025glm45v} & $24{,}363$ & $2{,}261$ & $1{,}242$ \\
InternVL3.5 (8B)~\citep{wang2025internvl35} & $13{,}892$ & $590$ & $1{,}245$ \\
InternVL3.5 (30B, A3B) & $5{,}461$ & $786$ & $622$ \\
InternVL3.5 (38B) & $62{,}411$ & $2{,}097$ & $1{,}556$ \\
MiMo-VL (7B)~\citep{xiaomi2025mimovl} & $12{,}759$ & $590$ & $1{,}243$ \\
MiniCPM-V 4.5 (9B)~\citep{yu2025minicpmv45} & $13{,}892$ & $590$ & $1{,}243$ \\
MiniCPM-V 2.6 (8B)~\citep{yao2024minicpmv} & $13{,}051$ & $401$ & $1{,}087$ \\
LongVILA (7B)~\citep{chen2024longvilascalinglongcontextvisual} & $13{,}051$ & $401$ & $1{,}087$ \\
Phi-4-MM (6B)~\citep{abouelenin2025phi4mini} & $7{,}181$ & $393$ & $1{,}229$ \\
\bottomrule
\end{tabular}
\caption{The numbers we use for FLOPs accounting, where $B$ is cost except for attention; $A$ is
what one query--key pair costs; $F_{\mathrm{voc}}$ is the output projection to the vocabulary distribution. }
\label{tab:flops_coeffs}
\end{table}

\paragraph{Checkpoint context lengths.} At $L{=} 1024$ for video, the context $N$ exceeds the largest
position the checkpoint was trained on for the three InternVL3.5 models
($6.4\times$), Phi-4-MM ($4.3\times$), MiniCPM-V 2.6 and GLM-4.5V
($2.0\times$), Qwen3-Omni ($1.5\times$), and marginally Qwen2.5-VL and
MiMo-VL ($1.02\times$). The models that assign positions along separate time and space axes
(the Qwen family, GLM-4.5V) advance their time index once per frame group, so
their largest position index is still below the token count, whereas the
InternVL and MiniCPM models index positions along a single axis and must
extrapolate heavily outside of their given context length.

\subsection{Per-model synthetic accuracy}\label{app:synth_numbers}
The figures in Section~\ref{sec:compression} plot accuracy against
compute, and Table~\ref{tab:synth_numbers} gives the same accuracies as numbers, including models omitted from the figures for space or for scoring below $51\,\%$ in every cell. Abstentions count as wrong.

\begin{table}[!htb]
\centering
\small
\begin{tabular}{l rr rr rr rr}
\toprule
& \multicolumn{2}{c}{text, $L{=}512$} & \multicolumn{2}{c}{text, $L{=}1024$} & \multicolumn{2}{c}{video, $L{=}16$} & \multicolumn{2}{c}{video, $L{=}64$} \\
\cmidrule(lr){2-3}\cmidrule(lr){4-5}\cmidrule(lr){6-7}\cmidrule(lr){8-9}
Model & low & max & low & max & low & max & low & max \\
\midrule
GLM-4.5V (106B, A12B) & 49.2 & 43.4 & 49.1 & 41.0 & 73.8 & 73.8 & 75.6 & 71.5 \\
InternVL3.5 (30B, A3B) & 96.7 & 75.7 & 94.3 & 67.9 & 53.4 & 53.1 & 48.4 & 48.4 \\
InternVL3.5 (38B) & 85.8 & 76.7 & 89.6 & 70.1 & -- & -- & -- & -- \\
InternVL3.5 (8B) & 32.8 & 54.0 & 23.6 & 41.0 & 69.7 & 74.1 & 50.0 & 48.4 \\
InternVL3.5 Thinking (30B, A3B) & 54.1 & 52.6 & 34.0 & 23.9 & 70.6 & 54.7 & 54.5 & 48.1 \\
InternVL3.5 Thinking (38B) & 74.9 & 51.0 & 67.9 & 26.1 & 52.8 & 54.7 & 57.5 & 53.8 \\
InternVL3.5 Thinking (8B) & 25.1 & 24.3 & 14.2 & 15.7 & 82.8 & 86.9 & 50.6 & 50.9 \\
LongVILA (7B) & 55.2 & 41.1 & 49.1 & 59.0 & 45.6 & 45.6 & 38.8 & 27.2 \\
MiMo-VL (7B) & 31.1 & 13.9 & 18.9 & 10.4 & 69.7 & 68.4 & 40.6 & 50.3 \\
MiniCPM-V 2.6 (8B) & 15.8 & 22.3 & 14.2 & 16.4 & 52.5 & 54.1 & 51.6 & 53.1 \\
MiniCPM-V 4.5 (9B) & 51.4 & 40.1 & 56.6 & 42.5 & 59.1 & 55.3 & 55.3 & 49.7 \\
Phi-4-MM (6B) & 78.7 & 54.0 & 71.7 & 58.2 & 44.7 & 40.0 & 46.9 & 37.2 \\
Qwen2.5-VL (7B) & 21.9 & 47.5 & 18.9 & 39.6 & 68.8 & 65.9 & 67.5 & 64.4 \\
Qwen3-Omni (30B, A3B) & 23.5 & 35.1 & 25.5 & 26.9 & 69.4 & 72.8 & 65.6 & 60.0 \\
Qwen3-VL (8B) & 85.8 & 71.3 & 84.0 & 60.4 & 61.9 & 63.4 & 59.4 & 60.0 \\
Qwen3-VL Thinking (8B) & 16.9 & 23.8 & 12.3 & 4.5 & 49.4 & 44.7 & 40.9 & 44.1 \\
\bottomrule
\end{tabular}
\caption{Per-model synthetic accuracy (\%) at the lengths plotted in Figures~\ref{fig:flops-1024-pretrained} and~\ref{fig:flops-video-64-sequential}, at both entropy levels. Abstentions count as wrong. InternVL3.5 (38B) was evaluated on synthetic video only at $L{=}8$ (Table~\ref{tab:short_L_perf}).}
\label{tab:synth_numbers}
\end{table}

\subsection{\texorpdfstring{Short-$L$ perception check}{Short-L perception check}}
\label{app:short_L_perception}

\textbf{Most models can read short streams, so their collapse at longer $L$ is a memory failure.} In order to make sure that the models have the basic skills required to perform our task, we conduct a short perception check. At $L{=}8$ little memory is needed to answer well, and the cohort of models we evaluate averages $86.8\,\%$ on text and $71.0$ and $65.1\,\%$ on the video variants (Table~\ref{tab:short_L_perf}). We interpret the model's $L{=}8$ score as its baseline and interpret the drop at longer lengths as degradation of its memory.

\begin{table}[!htb]
\centering
\small
\begin{tabular}{lccc}
\toprule
 & Text (\%) & Video (\%) & Video (\%) \\
Model & ($L{=}8$) & ($L{=}8$, sequential) & ($L{=}8$, spatial) \\
\midrule
Qwen3-Omni (30B, A3B) & 100.0 & 81.1 & 70.8 \\
MiniCPM-V 4.5 (9B) & 49.5 & 77.4 & 62.1 \\
MiniCPM-V 2.6 (8B) & 84.8 & 59.7 & 52.8 \\
LongVILA (7B) & 62.0 & 48.4 & 49.1 \\
Qwen2.5-VL (7B) & 16.3 & 66.0 & 67.1 \\
Qwen3-VL (8B) & 83.2 & 71.1 & 66.5 \\
Qwen3-VL Thinking (8B) & 100.0 & 50.3 & 42.2 \\
GLM-4.5V (106B, A12B) & 99.5 & 69.2 & 72.0 \\
InternVL3.5 (8B) & 100.0 & 91.8 & 49.7 \\
InternVL3.5 Thinking (8B) & 100.0 & 91.8 & 91.3 \\
Phi-4-MM (6B) & 98.9 & 45.3 & 42.2 \\
InternVL3.5 (30B, A3B) & 99.5 & 78.0 & 67.1 \\
InternVL3.5 Thinking (30B, A3B) & 100.0 & 86.8 & 84.5 \\
MiMo-VL (7B) & 96.7 & 75.5 & 73.3 \\
InternVL3.5 (38B) & 98.4 & 80.5 & 87.0 \\
InternVL3.5 Thinking (38B) & 100.0 & 62.9 & 64.0 \\
\midrule
\textit{Cohort mean} & 86.8 & 71.0 & 65.1 \\
\bottomrule
\end{tabular}
\caption{Raw accuracy (\%) at $L{=}8$ on the low-entropy band, for text and for both video question variants; abstentions count as wrong. Qwen2.5-VL's text score at $L{=}8$ comes from writing the yes and no markers swapped.}
\label{tab:short_L_perf}
\end{table}

\subsection{Natural-video evaluation}
\label{app:natural_eval}

\subsubsection{Construction}
\label{app:natural_construction}
The natural-video benchmark consists of binary membership questions over short clips drawn primarily from EPIC-Kitchens-100, with a smaller share from SoccerNet in the same yes/no/abstain answer format as the rest of ECCBench.

\paragraph{Frame extraction at 1\,fps.}
Every video is rendered as a sequence of frames which we extract at 1 fps. The frame count therefore equals the duration in seconds. Each model receives the full extracted frame sequence as input; we do not sub-sample down to a small number of frames---a common practice in long-video benchmarks where a ``long'' video is presented as 8, 16, or 32 uniformly sampled frames regardless of source length. Under that convention and unlike our evaluation, a model never really processes a long video. 

For most videos the source clip is at least as long as the target frame count in seconds. Some source clips are shorter than the target length, and in those cases we apply a slowdown to the source clip---extracting the same content at sub-real-time speed so that 1 frame per second still yields the target frame count. We record the slowdown per-video.

\paragraph{Length buckets.} We created six buckets, $L \in \{8,16,64,128,512,1024\}$ frames, holding 252 videos each at $L \in \{8,16,64\}$, 110 each at $L \in \{128,512\}$ and 52 at $L{=}1024$. Per-source composition: at $L\in\{8,16,64\}$, 236 of 252 videos per bucket are from EPIC-Kitchens-100 and 16 from SoccerNet; at $L\in\{128,512\}$, 100 of 110 are EPIC, 10 are SoccerNet; at $L{=}1024$, 42 of 52 are EPIC, 10 are SoccerNet.

\paragraph{Slowdown distribution.} $L{=}8$ and $L{=}16$ have no slowed-down videos. $L{=}64$ contains 2 out of 252 total. $L{=}128$ contains 3 of 110. $L{=}512$ contains 15 of 110. $L{=}1024$ contains 42 of 52.

\paragraph{Event categories.} Each video is associated with one of eleven event categories: \texttt{open\_fridge}, \texttt{open\_cupboard}, \texttt{open\_drawer}, \texttt{pour\_water}, \texttt{cut\_nontomato}, \texttt{rinse\_X}, \texttt{turn\_on\_tap}, \texttt{wipe\_surface}, \texttt{clean\_counter}, \texttt{wash\_plate}, and \texttt{red\_card}. The first ten are kitchen-action categories from EPIC-Kitchens-100 annotations, and \texttt{red\_card} is from SoccerNet.

\paragraph{Question format.} Each question takes the form \emph{``Did the person [action] in this video?''} for kitchen events, or \emph{``Did a red card appear in this video?''} for the SoccerNet category. The ground truth comes from the source annotations.

\paragraph{Clips and questions used in each experiment.} We compare models on a subset chosen to make visual density comparable across lengths: 163, 87, 74, and 36 clips at $L{=}64,128,512,1024$, respectively. Each clip has a question about its associated event, plus questions about other annotated events. Figure~\ref{fig:natvid_density} uses the same 354 questions at $L{=}64$ and 342 at $L{=}128$ for every method. Table~\ref{tab:band_contrast} uses these same questions at those lengths and model-specific subsets at longer lengths. The calibration experiment uses only the additional event questions, paired across its two payoff conditions. Table~\ref{tab:rag} compares RAG and the full-video model on the same 354, 342, and 199 questions at $L{=}64,128,512$.

Figure~\ref{fig:natvid_density} uses MiniCPM-o 2.6. Its dimensions follow the MiniCPM-V 2.6 row in Table~\ref{tab:flops_constants}, with a vocabulary of 151,700 tokens.

\subsubsection{Entropy bands}
\label{app:natural_density}
Natural footage has no known ground-truth entropy, so information density is measured by
fixing a coding \emph{rate} and reading off the resulting \emph{distortion} (Section~\ref{sec:data}). We use a cutoff to assign each video to
the low or the high band; every
question asked about a clip inherits this designation, similarly to our synthetic subset.

\paragraph{Encoding.} Each clip is re-encoded
with x264 at $32$\,kbit per frame, at $640\times360$, to a constant $1$\,fps. Rate
control is two-pass with no cap on the instantaneous rate, since single-pass
average-bitrate control does not converge on clips this short and a cap forces
undershoot. The keyframe interval exceeds every clip length, so no periodic intra frame
is inserted. We kept scene-cut detection enabled. The
preset is \texttt{veryslow} with up to $8$ B-frames and $16$ reference frames.

\paragraph{Distortion.} We compute and compare distortion in the embedding space of DINOv2~\citep{oquab2024dinov2}. With
$z_t$ and $\hat z_t$ the $\ell_2$-normalized patch-mean DINOv2
embeddings of the original and the decoded frame $t$, the per-frame distortion
is
\[
\delta_t \;=\; 1 - \langle z_t,\, \hat z_t \rangle ,
\]
and the clip score is the mean of $\delta_t$ over frames, dropping frame $0$,
which is coded from scratch.

\paragraph{Banding.} We define a clip as low-entropy if its distortion score is below a
cutoff of $0.0401$ and consider it high-entropy otherwise. Table~\ref{tab:codec_band_composition} shows the distortion statistics of our dataset.

\begin{table}[!htb]
\centering
\small
\begin{tabular}{lcccc}
\toprule
$L$ & Videos & Questions & Mean distortion & Yes-rate \\
    & (low / high) & (low / high) & (low / high) & (low / high) \\
\midrule
64   & 33 / 130 & 67 / 287  & 0.0306 / 0.0704 & 45\,\% / 57\,\% \\
128  & 28 / 59  & 107 / 235 & 0.0317 / 0.0613 & 46\,\% / 57\,\% \\
512  & 28 / 46  & 100 / 178 & 0.0303 / 0.0542 & 52\,\% / 62\,\% \\
1024 & 24 / 12  & 93 / 44   & 0.0302 / 0.0478 & 57\,\% / 66\,\% \\
\midrule
Total & & 367 / 744 & & \\
\bottomrule
\end{tabular}
\caption{Composition of the selected clip and question pool using the entropy cutoff of $0.0401$. The yes-rate is the percentage of questions with positive ground-truth
answers.}
\label{tab:codec_band_composition}
\end{table}

\textbf{The rationale behind our metric for compressibility.}
The metric should agree with intuitive notions of compressibility while extending to images that differ in pixel space but carry the same information for natural language questions. For instance, duplicating frames in a video should score as more compressible on average (relative to a video of the same length without duplicated frames). The comparison in an embedding space fits these desiderata.

\textbf{VLMs do not show the consistent ability to compress natural video.} We show the contrast, defined as the accuracy on the low-entropy split minus the high-entropy split, in Table~\ref{tab:band_contrast}. InternVL3.5 (8B) shows the largest contrast at both lengths in Figure~\ref{fig:natvid_density}. InternVL3.5 (8B) was evaluated at $L{=}64$ and $L{=}128$ only. The two bands also differ in label prior (Table~\ref{tab:codec_band_composition}), so we also score every cell with both answer classes weighted equally: the mean contrast moves from $-2.32$ to $-1.31$ points, positive in 10 of 41 cells. InternVL3.5 (8B)'s gap between low and high still survives at $+14.7$ at $L{=}128$. The $95\,\%$ confidence intervals show how much our accuracy results are robust and hold up to picking a random group of videos. To estimate them, we run a cluster bootstrap where for each model and length, we build $4{,}000$ mock test sets by randomly picking clips from each entropy band.

Slowdown can make neighboring frames more similar, making a clip more compressible. This is a confounding factor that correlates with, but perhaps is not entirely captured by, our distortion factor. We therefore repeat the comparison using only unslowed clips at $L\in\{64,128,512\}$. The contrast between low- and high-entropy accuracy is $-2.53$ points before removing slowed clips and $-2.38$ afterward, with a $95\,\%$ interval of $[-5.78,0.94]$. We can thus still conclude, with or without slowed videos, that there is little evidence that VLMs can consistently compress natural video.

\begin{table}[!htb]
\input{tables/table_band_contrast}
\caption{Each cell is accuracy on
the low-entropy band minus accuracy on the high-entropy band in percentage
points, with $95\,\%$  intervals underneath. A positive value means the model did better on
the low-entropy videos. About a quarter of the cells show a gain and most of the rest show a loss, and the mean over
all 41 cells is $-2.32$, indicating very little compressive capability, though there may be some present that would require a large test. $\dagger$ marks cells where the model
scores under $5\,\%$ on both bands; row means omit these cells. -- marks cells that were not evaluated.}
\label{tab:band_contrast}
\end{table}

\subsection{Calibration: the two payoff arms}
\label{app:calibration_arms}
\paragraph{Exact prompt modification.}
We append a new sentence to the original prompt in
Section~\ref{app:prompts} and do not otherwise change it. \texttt{\{X\}} is \texttt{\{uncertain\}} for natural video, \texttt{\{2\}} otherwise.

\begin{verbatim}
arm A:
Scoring: a correct answer earns +3 points, a wrong answer
costs -1 point, and {X} earns 0 points. Answer only if you
are more than 25% confident; otherwise answer {X}.

arm B:
Scoring: a correct answer earns +1 point, a wrong answer
costs -4 points, and {X} earns 0 points. Answer only if you
are more than 80% confident; otherwise answer {X}.
\end{verbatim}

\paragraph{Calibration arm scoring.} In Table~\ref{tab:calib_all_cells}, we show the runs behind Section~\ref{sec:calibration_main} and additional data from other models.

\begin{table}[!htb]
\centering
\small
\begin{tabular}{@{}l c r cc cc c@{}}
\toprule
& & & \multicolumn{2}{c}{coverage} & \multicolumn{2}{c}{acc.\ answered (\%)} & \\
\cmidrule(lr){4-5}\cmidrule(lr){6-7}
Model & $L$ & $n$ & arm A & arm B & arm A & arm B & gap \\
\midrule
\multicolumn{8}{@{}l}{\emph{synthetic text}} \\
InternVL3.5 30B & $512, 1024$ & $644$ & $1.000$ & $0.998$ & $81.8$ & $82.3$ & -- \\
InternVL3.5 38B & $512, 1024$ & $644$ & $0.992$ & $0.984$ & $86.1$ & $81.7$ & $+0.084$ \\
InternVL3.5 8B & $512, 1024$ & $487$ & $1.000$ & $1.000$ & $83.4$ & $80.3$ & -- \\
InternVL3.5 Thinking 30B & $512, 1024$ & $480$ & $0.590$ & $0.550$ & $81.6$ & $84.1$ & $+0.102$ \\
InternVL3.5 Thinking 38B & $512, 1024$ & $644$ & $0.620$ & $0.648$ & $90.5$ & $86.6$ & $+0.116$ \\
InternVL3.5 Thinking 8B & $512, 1024$ & $323$ & $0.337$ & $0.344$ & $71.6$ & $78.4$ & $+0.304$ \\
MiMo-VL & $512, 1024$ & $323$ & $0.260$ & $0.279$ & $60.7$ & $63.3$ & $+0.053$ \\
Phi-4-MM & $512, 1024$ & $644$ & $0.677$ & $0.702$ & $62.6$ & $62.8$ & $+0.045$ \\
Qwen3-Omni & $512, 1024$ & $644$ & $0.393$ & $0.474$ & $90.1$ & $87.9$ & $+0.178$ \\
Qwen3-VL & $512, 1024$ & $644$ & $0.995$ & $0.994$ & $70.4$ & $74.1$ & -- \\
Qwen3-VL Thinking & $512, 1024$ & $323$ & $0.226$ & $0.207$ & $76.7$ & $71.6$ & $+0.101$ \\
\midrule
\multicolumn{8}{@{}l}{\emph{synthetic video}} \\
InternVL3.5 30B & $8, 16$ & $640$ & $1.000$ & $1.000$ & $48.6$ & $48.6$ & -- \\
InternVL3.5 38B & $8, 16$ & $640$ & $1.000$ & $1.000$ & $48.6$ & $48.6$ & -- \\
InternVL3.5 8B & $8, 16$ & $1280$ & $1.000$ & $1.000$ & $58.0$ & $55.1$ & -- \\
InternVL3.5 Thinking 30B & $8, 16$ & $1280$ & $0.857$ & $0.884$ & $54.8$ & $55.2$ & $-0.029$ \\
InternVL3.5 Thinking 8B & $8, 16$ & $1280$ & $0.947$ & $0.959$ & $53.9$ & $54.8$ & $+0.105$ \\
MiMo-VL & $8, 16$ & $960$ & $0.982$ & $0.973$ & $72.1$ & $73.2$ & $+0.071$ \\
MiniCPM-V 4.5 & $8, 16$ & $1280$ & $0.981$ & $0.984$ & $67.5$ & $67.9$ & $-0.104$ \\
Qwen2.5-VL & $8, 16$ & $1280$ & $0.999$ & $0.977$ & $66.2$ & $67.1$ & $-0.309$ \\
Qwen3-Omni & $8, 16$ & $1280$ & $1.000$ & $1.000$ & $66.6$ & $68.0$ & -- \\
Qwen3-VL & $8, 16$ & $1280$ & $1.000$ & $1.000$ & $72.2$ & $69.2$ & -- \\
Qwen3-VL Thinking & $8, 16$ & $1280$ & $0.588$ & $0.566$ & $69.3$ & $70.9$ & $+0.145$ \\
\midrule
\multicolumn{8}{@{}l}{\emph{natural video}} \\
GLM-4.5V & $64$ & $176$ & $0.994$ & $1.000$ & $90.9$ & $90.3$ & -- \\
InternVL3.5 30B & $64$ & $176$ & $0.994$ & $1.000$ & $88.6$ & $87.5$ & -- \\
InternVL3.5 38B & $64$ & $176$ & $1.000$ & $1.000$ & $91.5$ & $91.5$ & -- \\
InternVL3.5 8B & $64$ & $176$ & $1.000$ & $1.000$ & $81.8$ & $81.8$ & -- \\
InternVL3.5 Thinking 30B & $64$ & $176$ & $0.989$ & $0.989$ & $86.2$ & $87.9$ & -- \\
InternVL3.5 Thinking 8B & $64$ & $176$ & $0.994$ & $1.000$ & $89.7$ & $85.8$ & -- \\
MiMo-VL & $64$ & $176$ & $0.926$ & $0.903$ & $73.0$ & $76.7$ & $+0.067$ \\
MiniCPM-V 2.6 & $64$ & $176$ & $0.909$ & $0.909$ & $85.0$ & $83.8$ & $+0.465$ \\
Qwen2.5-VL & $64$ & $176$ & $0.670$ & $0.608$ & $66.9$ & $73.8$ & $+0.560$ \\
Qwen3-Omni & $64$ & $176$ & $1.000$ & $1.000$ & $81.8$ & $81.2$ & -- \\
Qwen3-VL & $64$ & $176$ & $1.000$ & $1.000$ & $92.0$ & $93.8$ & -- \\
GLM-4.5V & $128$ & $247$ & $0.996$ & $1.000$ & $89.8$ & $89.9$ & -- \\
InternVL3.5 30B & $128$ & $247$ & $1.000$ & $1.000$ & $85.8$ & $85.4$ & -- \\
InternVL3.5 38B & $128$ & $247$ & $1.000$ & $1.000$ & $93.9$ & $93.5$ & -- \\
InternVL3.5 Thinking 30B & $128$ & $247$ & $0.996$ & $0.996$ & $81.3$ & $80.5$ & -- \\
InternVL3.5 Thinking 8B & $128$ & $247$ & $0.992$ & $0.984$ & $83.3$ & $77.8$ & -- \\
MiMo-VL & $128$ & $247$ & $0.927$ & $0.951$ & $76.0$ & $80.4$ & $-0.075$ \\
MiniCPM-V 2.6 & $128$ & $247$ & $0.798$ & $0.781$ & $83.2$ & $86.0$ & $+0.050$ \\
Qwen2.5-VL & $128$ & $247$ & $0.623$ & $0.595$ & $74.7$ & $75.5$ & $+0.524$ \\
Qwen3-Omni & $128$ & $247$ & $1.000$ & $1.000$ & $80.2$ & $80.2$ & -- \\
Qwen3-VL & $128$ & $247$ & $1.000$ & $1.000$ & $87.4$ & $87.9$ & -- \\
GLM-4.5V & $512$ & $199$ & $0.960$ & $0.980$ & $83.2$ & $81.5$ & -- \\
MiMo-VL & $512$ & $199$ & $0.879$ & $0.910$ & $73.7$ & $71.8$ & $-0.173$ \\
Qwen3-Omni & $512$ & $199$ & $1.000$ & $1.000$ & $74.9$ & $73.9$ & -- \\
Qwen3-VL & $512$ & $199$ & $1.000$ & $1.000$ & $85.4$ & $86.4$ & -- \\
\bottomrule
\end{tabular}
\caption{All calibration data. $n$ is the number of queries; coverage is the fraction that the model chose to answer under each arm, maxing out at $1.0$. Accuracy is defined here over answered items only. \emph{gap} is arm~A accuracy on answers arm~B kept minus accuracy on those that arm~B abstained on, shown only where arm~B declined to answer at least five questions that arm~A answered.}
\label{tab:calib_all_cells}
\end{table}

\paragraph{Abstention.}
We are evaluating model behavior, and thus count a lack of an answer as an abstention, but three distinct behaviors are counted as abstention.

\emph{(i) Output limit.} A response that reaches the 8,192 token limit without an answer counts as an abstention.

\emph{(ii) Format non-compliance or no answer.} If a model does not give an answer in the proper bracketed format, or gives no answer at all, we count it as an abstention.

\emph{(iii) Explicit abstention.} The model may abstain explicitly by outputting the format corresponding to that option.

\FloatBarrier

\section{Backbone Architecture Experimental Details}
\label{app:backbone_details}

\subsection{Configurations and training}
Table~\ref{tab:backbone_configs} lists the architectural settings for the two embedding widths used in the backbone evaluation.

Note that we also trained a Transformer with no positional encoding at all~\citep{kazemnejad2023nope,koecher2025nope}. It never learned the task at either width, staying at chance on the training distribution, so it is not reported. We instead use a Transformer with a small convolution to evaluate the effects of RoPE.

\begin{table}[t]
    \centering
    \small
    \begin{tabular}{@{}l ccc >{\raggedright\arraybackslash}p{0.3\textwidth} >{\raggedright\arraybackslash}p{2cm}@{}}
    \toprule
    & \multicolumn{3}{c}{shared parameters} & & \\
    \cmidrule(lr){2-4}
    & & FFN & & & \\
    Model & heads & width & conv & Architecture-specific & $D{=}128$ \\
    \midrule
    Compressive Transf.         & 4 & $4D$ & --- & block and memory length 128; compression $[8,4]$ at lengths $[32,32]$ & --- \\
    DeltaFormer                 & 4 & 256 & --- & chunked attention; qk-norm; $\text{rope}_\theta{=}10^{4}$ & --- \\
    DeltaNet                    & 4 & 256 & 4 & $\text{expand}_k{=}\text{expand}_v{=}1$; L2 qk-norm & --- \\
    Gated DeltaNet              & 4 & $4D$ & 4 & $\text{expand}_v{=}2$; gating; short-conv preprocessing & --- \\
    GLA                         & 4 & 256 & --- & $\text{expand}_k{=}0.5$ ($d_k{=}8$); $\text{expand}_v{=}1$ & --- \\
    Mamba-2                     & --- & --- & 4 & $d_{\text{state}}{=}128$; expand 2 & $d_{\text{state}}{=}256$ \\
    Memory Mosaic               & 4 & --- & --- & block size 256; 6 persistent memories of size 128 & persistent size 256 \\
    MoM                         & 4 & $4D$ & --- & 16 memories; top-$k$ 3; capacity 1.5; Gated DeltaNet experts & --- \\
    RetNet                      & 4 & 256 & 4 & $\text{expand}_k{=}1$; $\text{expand}_v{=}2$; output gate & --- \\
    RWKV-7                      & 4 & $4D$ & --- & --- & --- \\
    GRU                         & --- & --- & --- & hidden width equals embedding width & --- \\
    Transformer (RoPE)          & 4 & $D$ & --- & causal; rotary encoding & --- \\
    Transformer (NoPE\,+\,conv) & 4 & $D$ & 4 & no positional encoding; depthwise conv mixer & --- \\
    TTT-MLP                     & 4 & $D$ & --- & window 16; mini-batch 256; gate & mini-batch 128 \\
    \bottomrule
    \end{tabular}
    \caption{Backbone hyperparameters, evaluated on sequence lengths 2048 and 4096,
    beyond the 32--1024 training window. Each architecture is three layers deep, has
    dropout 0, and uses a head dimension of 16. Going from $D{=}64$ to $D{=}128$ therefore doubles the width and
    the head count; the last column lists the settings that change beyond
    that. A dash means the parameter does not apply to that architecture or is not counted. FFN width is the inner width of the feed-forward layer, and $D$ is the embedding width.}
    \label{tab:backbone_configs}
\end{table}

\paragraph{Training.}
We train three-layer versions of all architectures for up to thirty-six wall-clock hours. For each model, we take the best of four training-configuration runs and use that one for evaluation. To simulate a streaming memory backbone, where the model is prepared to answer at any time step, we train on sequence lengths 32--1024.

All jobs reuse the same synthetic membership dataset: 32 queries per sequence, 500{,}000 randomly generated sequences, and a vocabulary of 16 symbols. We reserve 5\,\% of examples for validation, train with cosine LR schedules (1\,\% warmup), bf16-mixed precision, and deterministic dataloading. Batch sizes range from 12 to 128 and learning rates from 2e-4 to 7e-4, chosen per architecture and width (the $D{=}128$ runs generally use 2.5e-4 and the $D{=}64$ runs 6e-4--7e-4); batch size for the largest configurations is set by GPU memory. We train for up to 125 epochs and stop once validation accuracy reaches 0.97. Per-run values for every evaluated checkpoint are released with the configurations.

\subsection{FLOPs formulas and per-architecture values}
The backbone accounting is kept separate from the VLM accounting, since these models have no vision tower and are compared only among themselves. Elementwise operations, including normalization, RoPE, activation, and softmax, are omitted under our matmul-FLOP convention.

\paragraph{Transformer.}

This derivation covers both transformer rows. Rotary positional encoding is applied elementwise and thus we do not count it under our convention, as stated before. Therefore, the RoPE Transformer and a no-encoding Transformer of the same shape have the same cost, whereas the width-4 causal depthwise convolution adds $8D$ FLOPs per token per layer, or $24DT$ over three layers. This is about $0.16\,\%$ of the total. 

Each layer contains causal multi-head attention~\citep{vaswani2017attention} followed by an $r{=}1$ MLP. Both use pre-normalization and residual connections. Each attention head has width $d_{h} = D/H$, and causal attention contains $P(T) = T(T+1)/2$ valid query--key pairs.

\begin{center}\small
\begin{tabular}{@{}l r@{}}
\toprule
Q, K, V, Out projections & $4 \times  (T\times D)\,(D\times D)  = 8TD^{2}$ \\
attention scores Q@K$^{\mathsf{T}}$ & $2P(T)D$ \\
weighted values A@V & $2P(T)D$ \\
MLP, r=1 & $2 \times  (T\times D)\,(D\times D)  = 4TD^{2}$ \\
\midrule
Per layer & $12TD^{2} + 4P(T)D = 12TD^{2} + 2DT(T+1)$ \\
\bottomrule
\end{tabular}
\end{center}

For $N = 3$ layers,

$F = 3\cdot [12TD^{2} + 2DT(T+1)] / 10^{6}$ MFLOPs.

\paragraph{GRU~\citep{cho2014gru}.}

Each layer is a GRU with hidden width $D$. At every token, it computes input and recurrent projections for the reset gate, update gate, and candidate state.

\begin{center}\small
\begin{tabular}{@{}l r@{}}
\toprule
Input projections for 3 gates & $(1\times D)(D\times 3D)\ \text{per token} = 6D^{2}$ \\
Recurrent projections for 3 gates & $(1\times D)(D\times 3D)\ \text{per token} = 6D^{2}$ \\
\midrule
Per layer & $12TD^{2}$ \\
\bottomrule
\end{tabular}
\end{center}

Therefore,

$F = 12NTD^{2} / 10^{6}$ MFLOPs,

with $N = 3$.

\paragraph{Mamba-2~\citep{dao2024mamba2}.}

Each layer is one Mamba-2 block with:

\begin{center}\small
\begin{tabular}{@{}l r@{}}
\toprule
 & $d_{inner}    = 2D$ \\
 & $d_{state}    = 2D$ \\
 & $p          = 64$ \\
 & $H          = d_{inner} / p$ \\
 & $d_{in}       = 8D + H$ \\
 & $\mathit{conv\_dim}   = 6D$ \\
\bottomrule
\end{tabular}
\end{center}

Here, $p = 64$ is the Mamba-2 SSM head width.

For recurrent inference:

\begin{center}\small
\begin{tabular}{@{}l r@{}}
\toprule
input projection & $2TD\cdot d_{in}$ \\
depthwise convolution, width 4 & $8T\cdot \mathit{conv\_dim}$ \\
state update & $2T\cdot d_{inner}\cdot d_{state}$ \\
state readout & $2T\cdot d_{inner}\cdot d_{state}$ \\
output projection & $2T\cdot d_{inner}\cdot D$ \\
\midrule
Per layer & $F_{layer} = T[2D\cdot d_{in} + 8\cdot \mathit{conv\_dim} + 20D^{2}]$ \\
\bottomrule
\end{tabular}
\end{center}

For three layers,

$F = 3T[2D\cdot d_{in} + 8\cdot \mathit{conv\_dim} + 20D^{2}] / 10^{6}$ MFLOPs.

\paragraph{GLA~\citep{yang2024gla}.}
GLA uses $d_{k} = 8$ and $d_{v} = 16$ per head, alongside a rank-16 decay operation.
\begin{center}\small
\begin{tabular}{@{}l r@{}}
\toprule
$q$ and $k$ linear layers to width $D/2$ & $2D^{2}$ \\
$v$, output gate, and final linear layers & $6D^{2}$ \\
Decay layer ($D \to 16 \to D/2$) & $48D$ \\
Memory state updates & $32D$ \\
Feed-forward layer (width $256$) & $1536D$ \\
\midrule
Per layer per token & $c = 8D^{2} + 1616D$ \\
\bottomrule
\end{tabular}
\end{center}
The total compute is $F = 3cT / 10^{6}$ MFLOPs.

\paragraph{RetNet~\citep{sun2023retnet}.}
RetNet uses $d_{k} = 16$ and $d_{v} = 32$ per head, storing a $16 \times 32$ memory grid per head.
\begin{center}\small
\begin{tabular}{@{}l r@{}}
\toprule
$q$ and $k$ linear layers & $4D^{2}$ \\
$v$ and output gate linear layers & $8D^{2}$ \\
Final linear layer & $4D^{2}$ \\
1D local convolutions (width 4) & $32D$ \\
Memory state updates & $128D$ \\
Feed-forward layer (width $256$) & $1536D$ \\
\midrule
Per layer per token & $c = 16D^{2} + 1696D$ \\
\bottomrule
\end{tabular}
\end{center}
The total compute is $F = 3cT / 10^{6}$ MFLOPs.

\paragraph{DeltaNet~\citep{schlag2021fastweight,yang2024deltanet}.}
DeltaNet reads its previous state using $v_{\mathrm{old}} = S_{t-1}^{\mathsf T} k_{t}$, calculates an update, and writes the output. Each head sets $d_{k} = d_{v} = 16$.
\begin{center}\small
\begin{tabular}{@{}l r@{}}
\toprule
$q$, $k$, and $v$ linear layers & $6D^{2}$ \\
Final linear layer & $2D^{2}$ \\
Learning rate projection ($D \to H$) & $D^{2}/8$ \\
Three 1D local convolutions (width 4) & $24D$ \\
Memory reading (step 1) & $32D$ \\
Memory writing (step 2) & $32D$ \\
Memory reading (step 3) & $32D$ \\
Feed-forward layer (width $256$) & $1536D$ \\
\midrule
Per layer per token & $c = 8.125D^{2} + 1656D$ \\
\bottomrule
\end{tabular}
\end{center}
The total compute is $F = 3cT / 10^{6}$ MFLOPs.

\paragraph{Gated DeltaNet~\citep{yang2025gateddeltanet}.}
Gated DeltaNet uses $d_{k} = 16$ and $d_{v} = 32$ per head with a $16 \times 32$ state grid. We omit the elementwise decay operations as they add zero matrix-multiplication FLOPs.
\begin{center}\small
\begin{tabular}{@{}l r@{}}
\toprule
$q$ and $k$ linear layers & $4D^{2}$ \\
$v$, output gate, and final linear layers & $12D^{2}$ \\
Decay and learning rate heads & $D^{2}/4$ \\
1D local convolutions (width 4) & $32D$ \\
Memory state updates & $192D$ \\
Feed-forward layer (width $4D$) & $24D^{2}$ \\
\midrule
Per layer per token & $c = 40.25D^{2} + 224D$ \\
\bottomrule
\end{tabular}
\end{center}
The total compute is $F = 3cT / 10^{6}$ MFLOPs.

\paragraph{RWKV-7~\citep{peng2025rwkv7}.}
RWKV-7 holds a $16 \times 16$ memory grid per head, using four separate matrix operations on its state per token.
\begin{center}\small
\begin{tabular}{@{}l r@{}}
\toprule
State times vector pass & $2Hf^{2}$ \\
State update pass & $2Hf^{2}$ \\
Value-key pass & $2Hf^{2}$ \\
Output readout pass & $2Hf^{2}$ \\
\midrule
Recurrent total ($f=16$) & $8Hf^{2} = 128D$ \\
\bottomrule
\end{tabular}
\end{center}

Combining all layers gives:
\begin{center}\small
\begin{tabular}{@{}l r@{}}
\toprule
$r$, $k$, $v$, and final linear layers & $8D^{2}$ \\
Decay low-rank layers ($D \to 64 \to D$) & $256D$ \\
Update low-rank layers ($D \to 64 \to D$) & $256D$ \\
Output gate low-rank layers ($D \to 128 \to D$) & $512D$ \\
Recurrent state passes & $128D$ \\
Extra state connection ($D \to 16 \to D$, layers 2--3) & $64D$ \\
Feed-forward layer ($D \to 4D \to D$) & $16D^{2}$ \\
\midrule
Layer 1 total & $c_{1} = 24D^{2} + 1152D$ \\
Layers 2--3 total & $c_{2} = 24D^{2} + 1216D$ \\
Three-layer total per token & $c_{\mathrm{tot}} = 72D^{2} + 3584D$ \\
\bottomrule
\end{tabular}
\end{center}
The total compute is $F = Tc_{\mathrm{tot}} / 10^{6}$ MFLOPs.

\paragraph{DeltaFormer~\citep{xu2025deltaformer}.}
We calculate DeltaFormer's total operations directly across all text tokens without splitting the attention process into sub-steps.
\begin{center}\small
\begin{tabular}{@{}l r@{}}
\toprule
$q$, $k$, $v$, and final linear layers & $8TD^{2}$ \\
Learning rate projection ($D \to H$) & $TD^{2}/8$ \\
Feed-forward layer (width $256$) & $1536TD$ \\
Causal attention matrix operations & $4DT^{2}$ \\
\midrule
Per layer & $F_{\mathrm{layer}} = T[(65/8)D^{2} + 1536D] + 4DT^{2}$ \\
\bottomrule
\end{tabular}
\end{center}
For three layers, the total compute is $F = 3 \cdot F_{\mathrm{layer}} / 10^{6}$ MFLOPs.

\paragraph{TTT-MLP~\citep{sun2025learninglearntesttime}.}
Each layer uses a local window attention mechanism, an adaptive fast-weight memory, and a standard feed-forward network. Each head stores a $16 \to 64 \to 16$ network inside its state. We denote the total count of valid attention token pairs across window length $W$ as $P_{W}(T) = \sum_{t=1}^{T} \min(t, W+1)$. When $T > W$, this evaluates to $P_{W}(T) = (W+1)T - W(W+1)/2$. Setting $W = f = 16$, we compute:

\begin{center}\small
\begin{tabular}{@{}l r@{}}
\toprule
Window attention linear layers & $8TD^{2}$ \\
Window attention matrix operations & $4D \cdot P_{W}(T)$ \\
Fast-weight memory linear layers & $10TD^{2}$ \\
Learning rate projection ($D \to H$) & $TD^{2}/8$ \\
Fast-weight memory updates & $896DT$ \\
Feed-forward layer ($D \to D \to D$) & $4TD^{2}$ \\
\midrule
Per layer & $F_{\mathrm{layer}} = T[22.125D^{2} + 896D] + 4D \cdot P_{W}(T)$ \\
\bottomrule
\end{tabular}
\end{center}
For three layers, the total compute is $F = 3 \cdot F_{\mathrm{layer}} / 10^{6}$ MFLOPs.

\paragraph{MoM~\citep{du2025mom}.}
Each token updates four memory paths: three routed paths and one shared path. Each memory path uses a Gated DeltaNet state with $d_{k} = d_{v} = 16$. Because the feature projections are shared across branches, we compute them once per token. We maintain separate convolution and memory states across all four paths:

\begin{center}\small
\begin{tabular}{@{}l r@{}}
\toprule
Routing layer ($D \to 16$) & $32D$ \\
Shared feature linear layers & $6D^{2}$ \\
Shared decay and learning rate heads & $D^{2}/4$ \\
Four active memory paths: & \\
\quad 1D local convolutions (width 4) & $96D$ \\
\quad Memory state updates & $384D$ \\
Output gate and final linear layers & $4D^{2}$ \\
Feed-forward layer (width $4D$) & $24D^{2}$ \\
\midrule
Per layer per token & $c = 34.25D^{2} + 512D$ \\
\bottomrule
\end{tabular}
\end{center}
The total compute is $F = 3cT / 10^{6}$ MFLOPs.

\paragraph{Memory Mosaic~\citep{zhang2025memorymosaics}.}
When reusing cached sequence states, we omit the exponential moving average operations as they add zero matrix-multiplication FLOPs. We count the full sequence attention operations directly:

\begin{center}\small
\begin{tabular}{@{}l r@{}}
\toprule
Context key, value, and output linear layers & $6TD^{2}$ \\
Persistent key and output linear layers & $4TD^{2}$ \\
Sequence attention matrix operations & $2DT(T-1)$ \\
Persistent memory reads (6 reads, width $2D$) & $48TD^{2}$ \\
\midrule
Per layer & $F_{\mathrm{layer}} = 58TD^{2} + 2DT(T-1)$ \\
\bottomrule
\end{tabular}
\end{center}
For three layers, the total compute is $F = 3 \cdot F_{\mathrm{layer}} / 10^{6}$ MFLOPs.

\paragraph{Compressive Transformer~\citep{rae2019compressive}.}
This architecture processes input sequences in blocks of 128 tokens. New tokens attend to other tokens in the same block as well as $m_{b}$ past saved memory slots. For a block $b$, the total number of attention pairs is $P_{b} = 128 \cdot m_{b} + 8256$.

The number of active memory slots $m_{b}$ follows a warm-up sequence: $m_{b} = 0, 128, 144, 160, 164, \dots, 193, 193, \dots$

\begin{center}\small
\begin{tabular}{@{}l r@{}}
\toprule
Token linear layers & $1024D^{2}$ \\
Feed-forward layer ($D \to 4D \to D$) & $2048D^{2}$ \\
Gated residual connections & $3072D^{2}$ \\
Attention matrix operations & $6D \cdot P_{b}$ \\
\midrule
Full block cost (before memory cleanup) & $6144D^{2} + 6D \cdot P_{b}$ \\
\bottomrule
\end{tabular}
\end{center}

We perform periodic memory cleanup steps at specific block milestones:
\begin{center}\small
\begin{tabular}{@{}l r@{}}
\toprule
Block 2: 8$\times$ compression and cached updates & $320D^{2}$ \\
Block 4: 4$\times$ compression and cached updates & $48D^{2}$ \\
Block 12: Long-term memory update and cached updates & $32D^{2} + 16D$ \\
\bottomrule
\end{tabular}
\end{center}
Any other token attends to 193 past memory slots and costs $48D^{2} + 1164D$.

For $n \ge 12$ full blocks followed by one token,

$F = 3\cdot [\sum_b(6144D^{2} + 6D\cdot P_{b}) + (n-1)\cdot 320D^{2} + (n-3)\cdot 48D^{2} + (n-11)\cdot (32D^{2} + 16D) + 48D^{2} + 1164D] / 10^{6}$ MFLOPs.

\begin{table}[H]
\centering
\small
\begin{tabular}{l rr rr}
\toprule
& \multicolumn{2}{c}{$D{=}64$, MFLOPs} & \multicolumn{2}{c}{$D{=}128$, MFLOPs} \\
\cmidrule(lr){2-3}\cmidrule(lr){4-5}
Model & $L{=}2048$ & $L{=}4096$ & $L{=}2048$ & $L{=}4096$ \\
\midrule
Compressive Transformer & 1817.09 & 3711.26 & 6188.52 & 12550.06 \\
DeltaFormer & 4033.22 & 14508.49 & 8475.58 & 29835.08 \\
DeltaNet & 856.06 & 1711.69 & 2121.26 & 4241.48 \\
GLA & 837.17 & 1673.94 & 2077.19 & 4153.37 \\
Gated DeltaNet & 1101.54 & 2202.55 & 4229.92 & 8457.78 \\
Mamba-2 & 926.87 & 1853.29 & 3669.71 & 7337.63 \\
Memory Mosaic & 3071.73 & 9363.97 & 9064.12 & 24567.84 \\
MoM & 1063.78 & 2127.03 & 3852.25 & 7702.62 \\
RWKV-7 & 1074.27 & 2148.01 & 3357.08 & 6712.52 \\
RetNet & 1070.07 & 2139.62 & 2945.84 & 5890.24 \\
GRU & 302.14 & 604.13 & 1208.55 & 2416.51 \\
TTT-MLP & 936.21 & 1872.06 & 2986.54 & 5971.84 \\
Transformer (RoPE) & 1915.11 & 7051.30 & 4434.49 & 15310.85 \\
Transformer (conv) & 1918.26 & 7057.59 & 4440.79 & 15323.44 \\
\bottomrule
\end{tabular}
\caption{Backbone model computational complexity across embedding dimensions ($T=L+1$). These are the values used for the Pareto frontiers in
Figures~\ref{fig:backbone_calib_main} and~\ref{fig:covacc_entropy_4096}.}
\label{mflops-simp}

\end{table}

\subsection{Calibration: coverage vs.\ accuracy}
\label{app:backbone_plots}
For a fixed MFLOPs budget, we read increased (i) capacity as shifting the whole curve up and to the right, (ii) compression as a larger low- vs.\ high-entropy gap at the FLOPs budget, and (iii) calibration as better tradeoffs between coverage and accuracy.

\begin{figure}[!htb]
    \centering
    \includegraphics[width=\textwidth]{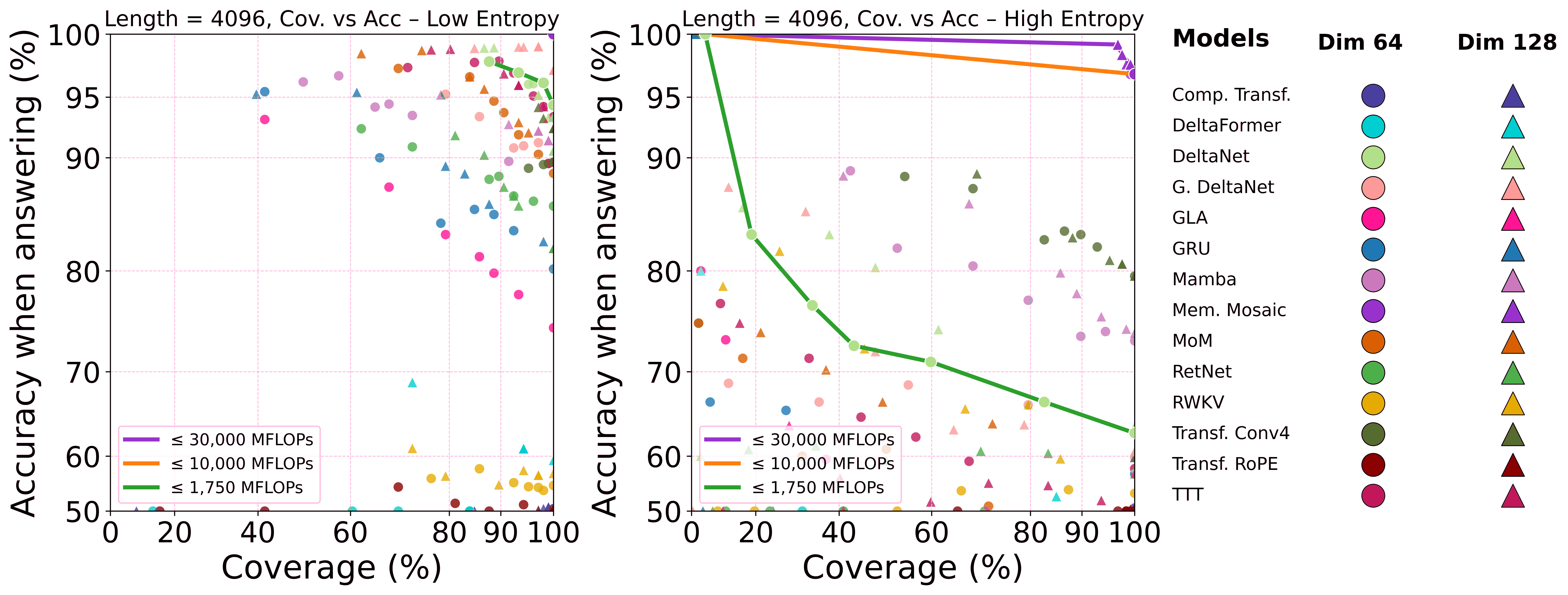}
    \caption{Same coverage-vs-accuracy convention and legend abbreviations as Figure~\ref{fig:backbone_calib_main}, now at $L{=}4096$. Both panels use FLOPs budgets $\le$30{,}000, $\le$10{,}000, $\le$1{,}750 MFLOPs (left: low-entropy split; right: high-entropy split). The $\le$30{,}000 tier admits all configurations.}
    \label{fig:covacc_entropy_4096}
\end{figure}

\FloatBarrier

\clearpage
\input{appendix_memory_methods.tex}

\section{Limitations}
\label{app:limitations_broader_impact}

\paragraph{Limitations.}
Our study codifies a general test of memory, but we do not test all types of questions that cover memorization. We omit natural text data, and while we do use natural video data, we draw the bulk of our conclusions from synthetic data. ECCBench isolates associative recognition and not the full range of behaviors commonly called memory. Our goal is not to test confounding skills that often impact tasks where memory is used, such as temporal reasoning, interference between different episodes or experiences, or whether remembered information can subsequently help complex tasks. ECC is intended to apply to the memory used in these settings, but we do not test on them.

We measure efficiency using arithmetic FLOPs and assume one query per stream. FLOPs provide an implementation-independent approximation to computation, but there are many other qualities that matter like memory bandwidth and communication costs. Moreover, systems that amortize an expensive write over many subsequent queries may rank differently under a multi-query workload.

Our synthetic experiments are designed to remove other skills and priors because what must be retained is known exactly, but they still rely on the ability of the model to extract and retain that information effectively. Thus, it is possible some models are able to store certain glyphs more compactly due to their priors, which would inflate their capacity score.

Our compressibility measure for natural video is a heuristic, validated only on sanity checks such as slowed-down clips scoring as more compressible. We chose it because no established measure of compressibility in feature space, rather than pixel space, exists.

We do not report complete results for closed-source models because we cannot access how many FLOPs they use. However, our definitions and axes of memory still apply to them.

Our backbone study uses small models, which limits the ability to generalize to LLM-scale, where models have billions of parameters. At that scale, it is possible that other capabilities emerge, and that the differences we observe are more or less pronounced. Some backbones may benefit from a wider hyperparameter search than the four configurations we tried. Nonetheless, as many studies of small models do generalize to larger ones, we believe it will be useful in motivating more efficient memory.

\section*{Acknowledgments}
This project was partially supported by the National Science Foundation.

%% file: tables/table_band_contrast.tex
\centering
\small
\begin{tabular}{lccccc}
\toprule
& \multicolumn{4}{c}{contrast at $L$} & \\
\cmidrule(lr){2-5}
Model & $64$ & $128$ & $512$ & $1024$ & mean \\
\midrule
GLM-4.5V & $-6.2$ & $-3.7$ & $+9.9$ & $+6.3$ & $+1.6$ \\
(106B, A12B) & $[-14.8, +1.9]$ & $[-11.8, +4.0]$ & $[+1.6, +18.2]$ & $[-7.4, +20.6]$ & \\
\addlinespace
InternVL3.5 & $-9.9$ & $-3.8$ & $+0.0^{\dagger}$ & $+0.0^{\dagger}$ & $-6.9$ \\
(30B, A3B) & $[-19.5, -0.9]$ & $[-12.8, +4.3]$ & $[+0.0, +0.0]$ & $[+0.0, +0.0]$ & \\
\addlinespace
InternVL3.5 & $-8.8$ & $-4.0$ & $+19.6$ & $+5.3$ & $+3.0$ \\
(38B) & $[-16.8, -0.7]$ & $[-11.3, +2.6]$ & $[+6.5, +33.4]$ & $[+0.0, +15.8]$ & \\
\addlinespace
InternVL3.5 & $+5.9$ & $+14.9$ & -- & -- & $+10.4$ \\
(8B) & $[-5.9, +16.9]$ & $[+2.9, +26.4]$ &  &  & \\
\addlinespace
InternVL3.5 Thinking & $-4.9$ & $-13.4$ & $+15.3$ & -- & $-1.0$ \\
(30B, A3B) & $[-14.9, +4.9]$ & $[-22.6, -4.5]$ & $[-5.6, +35.7]$ &  & \\
\addlinespace
InternVL3.5 Thinking & $-0.2$ & $-1.9$ & $+0.0^{\dagger}$ & $-3.3^{\dagger}$ & $-1.0$ \\
(8B) & $[-10.8, +9.7]$ & $[-12.0, +8.3]$ & $[+0.0, +0.0]$ & $[-10.0, +0.0]$ & \\
\addlinespace
MiMo-VL & $+0.6$ & $+9.1$ & $-1.3$ & $-4.7$ & $+0.9$ \\
(7B) & $[-10.6, +12.2]$ & $[-0.2, +18.0]$ & $[-11.5, +8.7]$ & $[-20.2, +12.4]$ & \\
\addlinespace
MiniCPM-o 2.6 & $-15.5$ & $-8.1$ & $-8.1$ & $+10.5$ & $-5.3$ \\
(8B) & $[-29.9, -2.0]$ & $[-17.0, +0.8]$ & $[-25.2, +8.9]$ & $[-2.7, +23.7]$ & \\
\addlinespace
Qwen2.5-VL & $-8.1$ & $-14.4$ & $-9.4$ & $-8.0$ & $-10.0$ \\
(7B) & $[-20.1, +2.9]$ & $[-28.1, -1.7]$ & $[-20.1, +1.9]$ & $[-24.5, +9.6]$ & \\
\addlinespace
Qwen3-Omni & $-4.2$ & $-12.3$ & $-5.7$ & $-14.3$ & $-9.1$ \\
(30B, A3B) & $[-14.4, +5.8]$ & $[-22.1, -2.6]$ & $[-16.5, +5.2]$ & $[-24.2, -4.4]$ & \\
\addlinespace
Qwen3-VL & $-4.2$ & $-2.9$ & $-5.5$ & $-5.8$ & $-4.6$ \\
(8B) & $[-12.7, +4.3]$ & $[-10.6, +4.7]$ & $[-15.0, +3.3]$ & $[-16.7, +4.9]$ & \\
\bottomrule
\end{tabular}

%% file: appendix_memory_methods.tex
\section{Other Memory Methods}
\label{app:other_memory_methods}

\subsection{RAG-Pipeline for Video Retrieval}\label{app:rag}

Our goal for ECC is to provide a general framework that can apply to many different types of systems. We therefore also evaluate our system on a RAG-like model, with an embedding and search procedure. Each video is split into 16-frame windows at a stride of 4 and every
such windowed clip is embedded with Qwen3-VL-Embedding-8B. When a query is received, the
candidate clip is embedded with the same model. We then rank it against all windows, and the top-ranked window (the top three at
$L{\geq}128$) is retrieved. This is then sent to Qwen3-Omni (30B, A3B) with the same prompts as every
baseline, plus a sentence saying that the clip was retrieved to provide some context for calibration purposes. 

The embedder shares the Qwen3-VL (8B) architecture, so we reuse those constants to calculate its prefill cost.
Embedding a query costs $0.56$ trillion on natural video, where the query is a
sentence, and $11$ trillion on synthetic video, where the query is
a video clip. Qwen3-Omni then answers as if on a clip of the retrieved length exactly as in Section~\ref{app:flops_per_model}.

\begin{table}[!htb]
\centering
\small
\begin{tabular}{l cc cc}
\toprule
& \multicolumn{2}{c}{accuracy (\%)} & \multicolumn{2}{c}{one query (TF)} \\
\cmidrule(lr){2-3}\cmidrule(lr){4-5}
corpus & RAG & full video & RAG & full video \\
\midrule
natural $L{=}64$   & $82.2$ & $82.5$ & $447$ & $87$ \\
natural $L{=}128$  & $81.0$ & $80.4$ & $994$ & $213$ \\
natural $L{=}512$  & $73.9$ & $71.9$ & $4{,}154$ & $1{,}786$ \\
synthetic $L{=}64$ & $58.4$ & $62.8$ & $413$ & $84$ \\
\bottomrule
\end{tabular}
\caption{The RAG-like pipeline versus Qwen3-Omni when it receives the full video, answering the same questions. RAG costs include building the index, embedding the query, and answering from the retrieved clip.}
\label{tab:rag}
\end{table}

For $q$ questions about the same stream, a comparison that reuses both systems' preparation gives average costs per answer
\[
\begin{aligned}
\overline F_{\mathrm{RAG}}(q) &= W_{\mathrm{RAG}}/q + R_{\mathrm{RAG}},\\
\overline F_{\mathrm{cached\ VLM}}(q) &= W_{\mathrm{VLM}}/q + R_{\mathrm{VLM}}.
\end{aligned}
\]
Here $W$ is the one-time preparation cost and $R$ the cost of answering each question. This assumes the VLM can retain and reuse its processed stream. Spreading preparation costs over questions does not necessarily make RAG cheaper than this alternative; generally, it will be cheaper when the embedding model is smaller or more efficient than the model that directly answers the questions.

Table~\ref{tab:rag} shows three main results. On natural video, the retrieval pipeline scores within one point of the full-video baseline at $L{=}64$ and $128$, and higher at $L{=}512$, but costs more when each answer includes building the index. On synthetic video, the pipeline scores $4.4$ points below the full-video baseline and answers only $37.0\,\%$ of yes-questions correctly. We also ran a retrieval probe to see how often the RAG model retrieves the right window; it succeeded only $33\,\%$ of the time in this setting. On natural video, no-questions are harder because the event is absent: the retriever must still return its closest match, and the model rarely rejects that window as irrelevant, causing no-question accuracy to fall from $62.9\,\%$ at $L{=}64$ to $46.3\,\%$ at $L{=}512$. Intuitively, retrieval also makes calibration harder because even weak matches are presented as evidence of a match (and the embedder never abstains). We theorize that providing retrieval similarity or confidence scores, as well as other metadata about the sequence, might help the model judge the quality of a match and infer what else is contained in the source video.

\subsection{Evaluating Other Memory Methods for VLMs}
\label{app:memory_methods}

Our evaluation is also designed to help in testing the impact of other modifications to memory systems, such as those designed to improve memory or save compute costs. We therefore use ECCBench to test recall, the benefit from regularity,
and calibration at a given FLOPs budget for SnapKV~\citep{li2024snapkv}
and InfLLM~\citep{xiao2024infllm}, which reduce attention to past tokens without
requiring retraining the model.
We test Mistral-7B-Instruct-v0.2 on text questions of $L{=}512$ and $L{=}1024$ symbols, and InternVL3.5-8B on $L{=}16$ synthetic video frames.
Our full control keeps all past tokens available. We use greedy
decoding and output at most $8{,}192$ tokens per question.

\paragraph{Using the question to choose memory.}

For InfLLM, we keep the context in $128$-token blocks and retrieve four or eight blocks
that best match the tokens it is processing, including the question. Each block
is represented by its four most-attended positions per head. It also keeps
$128$ initial tokens for the initial prompt, and a recent window of $512$ tokens for text or $2048$
for video (for the candidate). 

SnapKV instead selects once per question, using
attention from the last $32$ positions and smoothing each position's score over five neighboring
positions. For text, we use ordinary SnapKV with a $1024$-position cache per head, which
can also remove question tokens. For video, we select $1024$ stream token positions per head
and retain the entire candidate and question. We likewise retain the candidate
and question throughout answering with video InfLLM, expanding its recent
window when necessary.

\paragraph{Choosing memory before seeing the question.}
We also adapt SnapKV to select $1024$ positions from the stream before adding
the question. This discards information without knowing what will be asked and will be expected to lose considerable accuracy, especially on incompressible data. Repetition can
make that loss less harmful.

\begin{figure}[tbp]
\centering
\includegraphics[width=\textwidth]{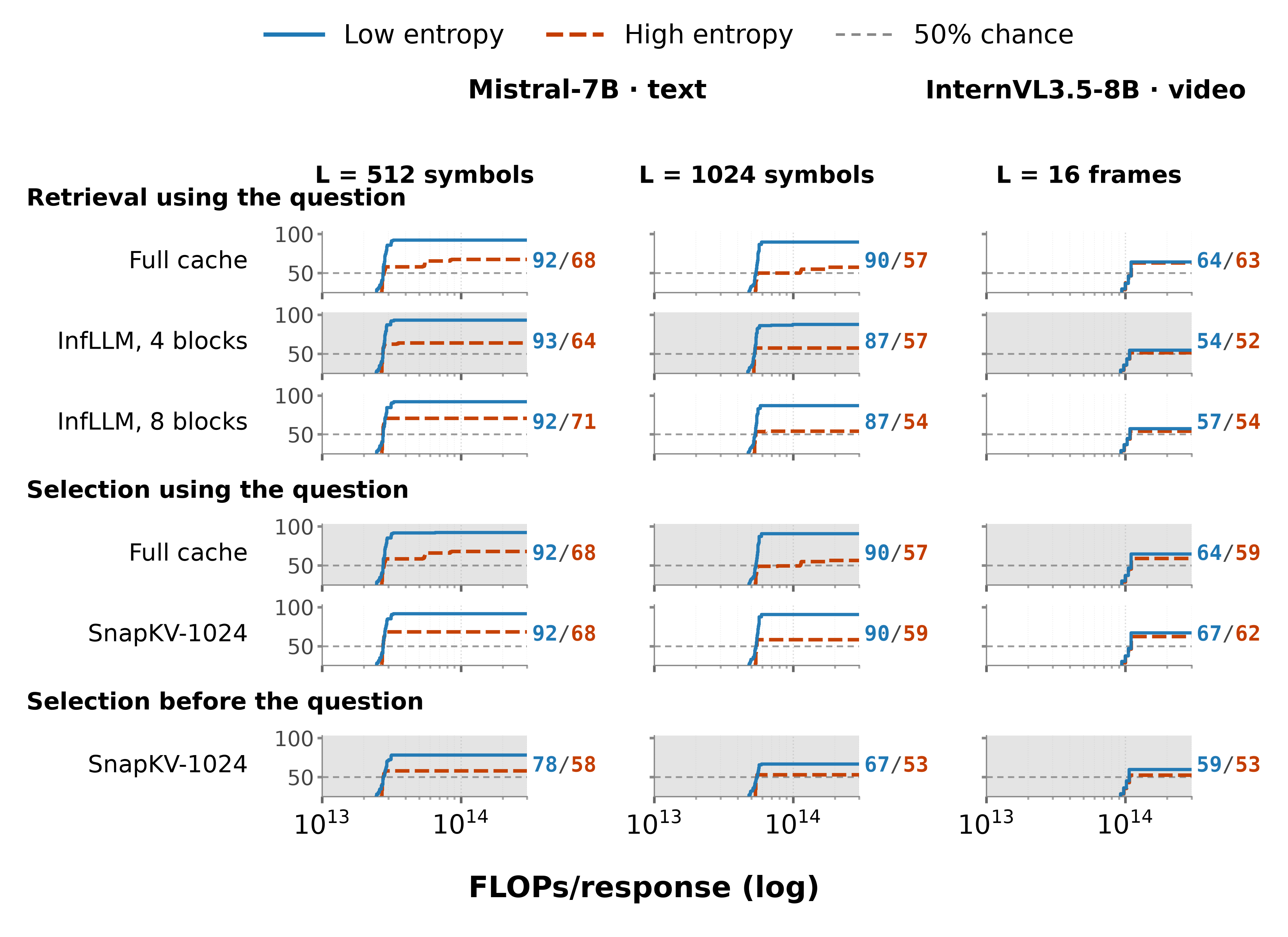}
\caption{Corresponding figure to Figure~\ref{fig:flops-1024-pretrained}. Each curve shows the percentage of all questions answered correctly within the budget including scoring and other preparation costs. Blue solid and orange dashed curves show low and maximal entropy. Numbers at the right give final accuracy.}
\label{fig:memory_methods_extended}
\end{figure}

\paragraph{Accuracy, cost, and regularity.}
When SnapKV sees the question first, Mistral scores $77.2\,\%$ against
$76.9\,\%$ for its full-cache control while saving $7.3\,\%$ of mean FLOPs.
InfLLM with four blocks scores $75.3\,\%$ against $76.7\,\%$, saving
$9.7\,\%$. If we allow the model to select the appropriate tokens before the question with SnapKV-1024 it saves $9.9\,\%$ of FLOPs,
but accuracy falls to $64.1\,\%$. At $L{=}1024$, regular streams score $32.5$ points above random streams with
full memory, compared with $13.8$ points with this version of SnapKV, showing worse compression ability
(Figure~\ref{fig:memory_methods_extended}). On video, SnapKV after the question
scores $64.7\,\%$ against $61.7\,\%$, but costs $0.05\,\%$ more. We suspect selection helps focus attention on relevant video tokens. InfLLM with
four blocks saves $2.4\,\%$ of FLOPs, but its accuracy falls significantly from $63.4\,\%$ to $53.0\,\%$.

\begin{figure}[tbp]
\centering
\includegraphics[width=\textwidth]{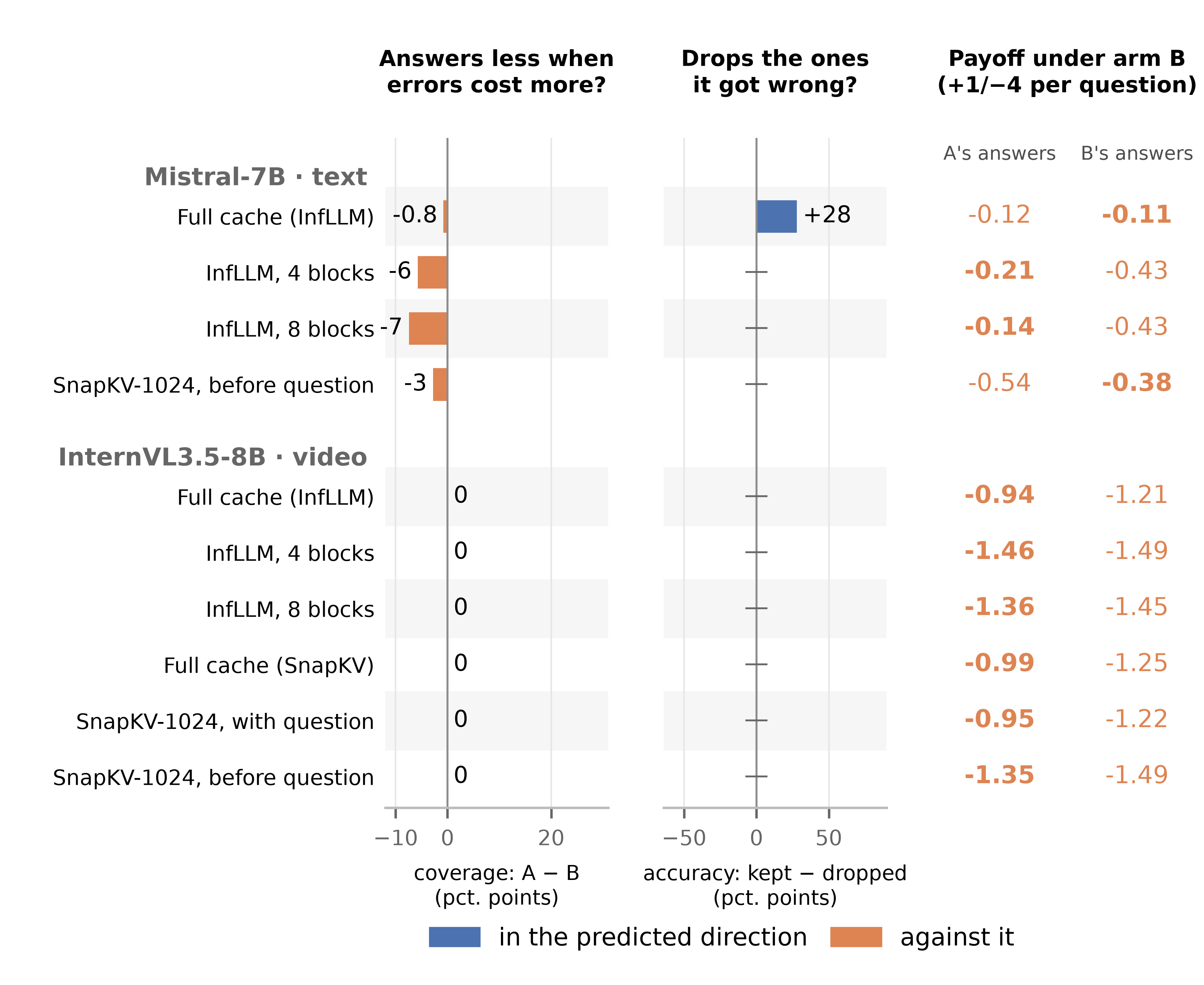}
\caption{Calibration as in Figure~\ref{fig:calib_resp_sel}. Left: coverage under A ($+3/-1$) minus B ($+1/-4$). Middle: difference in A accuracy between answers B keeps and drops; --- if either group is empty or fewer than five answers are dropped. Right: points earned under B's rule, with the higher value bold.}
\label{fig:memory_methods_calibration}
\end{figure}

\paragraph{Calibration under changing error costs.}
We then evaluate the calibration properties of these methods. When errors
cost more (under arm B), Mistral answers more often with SnapKV-1024 selecting before the
question and InfLLM using four blocks, with larger changes than the full-cache
control. This does not consistently help: SnapKV earns more points and InfLLM earns fewer, but both
remain below full memory and the option of always abstaining on each question. InternVL answers every question
under both rules with all methods, so neither improves abstention; both earn less than always abstaining under B (Figure~\ref{fig:memory_methods_calibration}).

\paragraph{Counting computation.}
We account for FLOPs for these methods following Appendix~\ref{app:flops_methodology}. SnapKV reads the full stream
before selecting positions. Selection after the question reduces attention
only during generation, because prefill still reads the whole stream. InfLLM restricts attention during stream reading and answering. We include the extra attention scores used by SnapKV and the
block-comparison scores used by InfLLM in the FLOPs calculations. 

For $D$ decoder layers, width
$d$, feed-forward width $f$, $H$ query heads, $K$ key/value heads, head width
$h$, and vocabulary size $V$, the per-token projections and feed-forward cost
per layer is $C=4dHh+4dKh+6df$. With $N$ input tokens and $T$ charged output
tokens, the total is
\[
F=F_{\mathrm{vision}}+DC(N+T-1)+4h\sum_{\ell=1}^{D}A_\ell+2dVT+S.
\]
Here $A_\ell$ counts attention pairs across all heads during reading and generation. The input pass predicts the first output token.
For video, $F_{\mathrm{vision}}$ follows Appendix~\ref{app:flops_per_model}.

For SnapKV, scoring $n$ tokens with a window of $w=32$ adds
$S=2DHh\{w(n-w)+w(w+1)/2\}$ when selection is necessary. For InfLLM,
scoring $g$ candidate blocks costs $2Hhg$ per layer per retrieval.
$S$ is the sum of these costs. When all available blocks fit the retrieval budget, block scoring is unnecessary.
For RAG, scoring $M$ stored window vectors of length $d_e$ costs $2Md_e$ FLOPs per question. Across all three methods, we count scoring products but exclude softmax, normalization, sorting, and transfers.

\FloatBarrier